\documentclass[11pt]{article}

\usepackage[final]{acl}

\usepackage{times}
\usepackage{latexsym}
\usepackage[T1]{fontenc}
\usepackage[utf8]{inputenc}
\usepackage{microtype}
\usepackage{inconsolata}
\usepackage{graphicx}
\usepackage{algorithm}
\usepackage{algorithmic}
\usepackage{amsmath}
\usepackage{amsfonts}
\usepackage{multirow}
\usepackage{diagbox}
\usepackage{subcaption}
\usepackage{balance}
\usepackage{tabularx}
\usepackage{array}
\usepackage{arydshln}

\usepackage{tcolorbox}
\usepackage{xcolor}
\definecolor{mygreen}{RGB}{0, 150, 0}
\definecolor{myred}{RGB}{200, 0, 0}

\usepackage{booktabs}   
\usepackage{colortbl}
\definecolor{graybg}{gray}{0.95}

\title{KARE-RAG: Knowledge-Aware Refinement and Enhancement for RAG}
\author{
  \normalfont
  Yongjian Li\textsuperscript{1,2,*},
  HaoCheng Chu\textsuperscript{3,*},
  Yukun Yan\textsuperscript{4,\textdagger},
  Zhenghao Liu\textsuperscript{8},
\\
  Shi Yu\textsuperscript{4},
  Zheni Zeng\textsuperscript{4},
  Ruobing Wang\textsuperscript{9},
  Sen Song\textsuperscript{1,2,\textdagger},
\\
  Zhiyuan Liu\textsuperscript{4,5,6,7},
  Maosong Sun\textsuperscript{4,5,6,7}
\\[-2pt]
  {\scriptsize \textsuperscript{1}School of Biomedical Engineering, Tsinghua Medicine, Tsinghua University}
\\[-2pt]
  {\scriptsize \textsuperscript{2}Tsinghua Laboratory of Brain and Intelligence \quad
  \textsuperscript{3}School of Informatics, Xiamen University}
\\[-2pt]
  {\scriptsize \textsuperscript{4}Department of Computer Science and Technology, Tsinghua University, Beijing}
\\[-2pt]
  {\scriptsize \textsuperscript{5}Beijing National Research Center for Information Science and Technology, Tsinghua University, Beijing}
\\[-2pt]
  {\scriptsize \textsuperscript{6}Institute for Artificial Intelligence, Tsinghua University, Beijing}
\\[-2pt]
  {\scriptsize \textsuperscript{7}Jiangsu Collaborative Innovation Center for Language Ability, Jiangsu Normal University, Xuzhou}
\\[-2pt]
  {\scriptsize \textsuperscript{8}School of Computer Science and Engineering, Northeastern University \quad
  \textsuperscript{9}Institute of Information Engineering, Chinese Academy of Sciences}
\\[-2pt]
  {\footnotesize \textsuperscript{*}Equal contribution. \quad \textsuperscript{\textdagger}Corresponding author. \quad \texttt{yj-li23@mails.tsinghua.edu.cn}}
}

\begin{document}
\maketitle

\begin{abstract}
Retrieval-Augmented Generation (RAG) equips large language models with external knowledge and is central to knowledge-intensive tasks. As RAG systems enter real-world use, generators must reliably leverage retrieved evidence. Recent fine-tuning methods improve adaptation to RAG scenarios, but optimization remains challenging because retrieval may return incomplete, fragmented, noisy, or conflicting contexts. Complex tasks further require fine-grained evidence dependencies. These challenges make high-quality supervision costly and limit generalization.

We present KARE-RAG (Knowledge-Aware Refinement and Enhancement for RAG), a training-time scaffolded alignment framework. It uses structured knowledge representations as temporary scaffolds to expose evidence organization, support localized factual refinement, and construct fine-grained preference pairs. In our main implementation, an expert LLM refines a lightweight graph-structured evidence sketch. The generator is optimized with token-weighted Dense Direct Preference Optimization (DDPO), which focuses learning on edited scaffold regions. Scaffolds are used only for data construction and training supervision. At inference time, the model runs as standard Vanilla RAG without graph construction, extra retrieval, or latency overhead. Experiments show that KARE improves transfer across the evaluated QA and relation extraction datasets with limited training data, while leaving general capabilities largely unchanged. KARE can also complement existing RAG training objectives as an additional alignment stage.
\end{abstract}

\section{Introduction}
Retrieval-Augmented Generation (RAG) \cite{lewis2020retrieval,shi2023replug} has become a cornerstone of knowledge-intensive tasks, enabling large language models (LLMs) to combine parametric knowledge with real-time information retrieval. Despite its widespread adoption, RAG systems face inherent challenges in processing retrieved documents. Evidence is often fragmented across multiple sources, requiring models to piece together complete information from partial segments \cite{xu2024search}. Retrieved documents may also present irrelevant or conflicting claims that require careful reconciliation \cite{wu2024easily,gao2023retrieval,yoran2023making,xu2024activerag,longpre2022entitybasedknowledgeconflictsquestion,liu2024lost}, as even state-of-the-art retrieval systems cannot guarantee perfect relevance in returned documents.

Existing research has made significant progress by improving or adapting retrieval strategies \cite{jiang2023active,gao2023retrieval,trivedi2022interleaving,mao2024rafe,ma2023query,jeong2024adaptive} or developing post-retrieval evidence processing mechanisms \cite{yu-etal-2024-chain,xu2023recomp,fang2024trace}. Recent studies show that improving how the generation module processes and integrates multi-source knowledge is an equally important, complementary approach \cite{singh2021end,lin2023ra,asai2023self,li2024rag}.

A common strategy is to fine-tune the model on RAG-style instruction data or final-answer supervision \cite{lin2023ra,asai2023self}. RA-DiT, for example, adapts RAG through dual instruction tuning and improves how the generator uses retrieved contexts \cite{lin2023ra}. However, answer-level supervised fine-tuning (SFT) provides only coarse supervision for complex evidence use and may overfit to the training distribution. Preference-based alignment offers another route. DDR constructs metric-derived preference pairs and applies Direct Preference Optimization (DPO) \cite{rafailov2024direct} to optimize generation under retrieved contexts \cite{li2024rag}. Such methods improve end-to-end behavior, but their supervision still falls mainly on final outputs. They provide limited guidance for intermediate evidence filtering, organization, and integration, where errors often arise in multi-hop reasoning or conflict resolution. Moreover, with intricate queries and noisy retrieval, sampling reliable positive examples can be difficult. This often requires large amounts of high-quality training data \cite{li2024rag}, motivating more targeted supervision for how RAG generators process evidence.

\begin{figure*}
    \centering
    \includegraphics[width=0.95\linewidth]{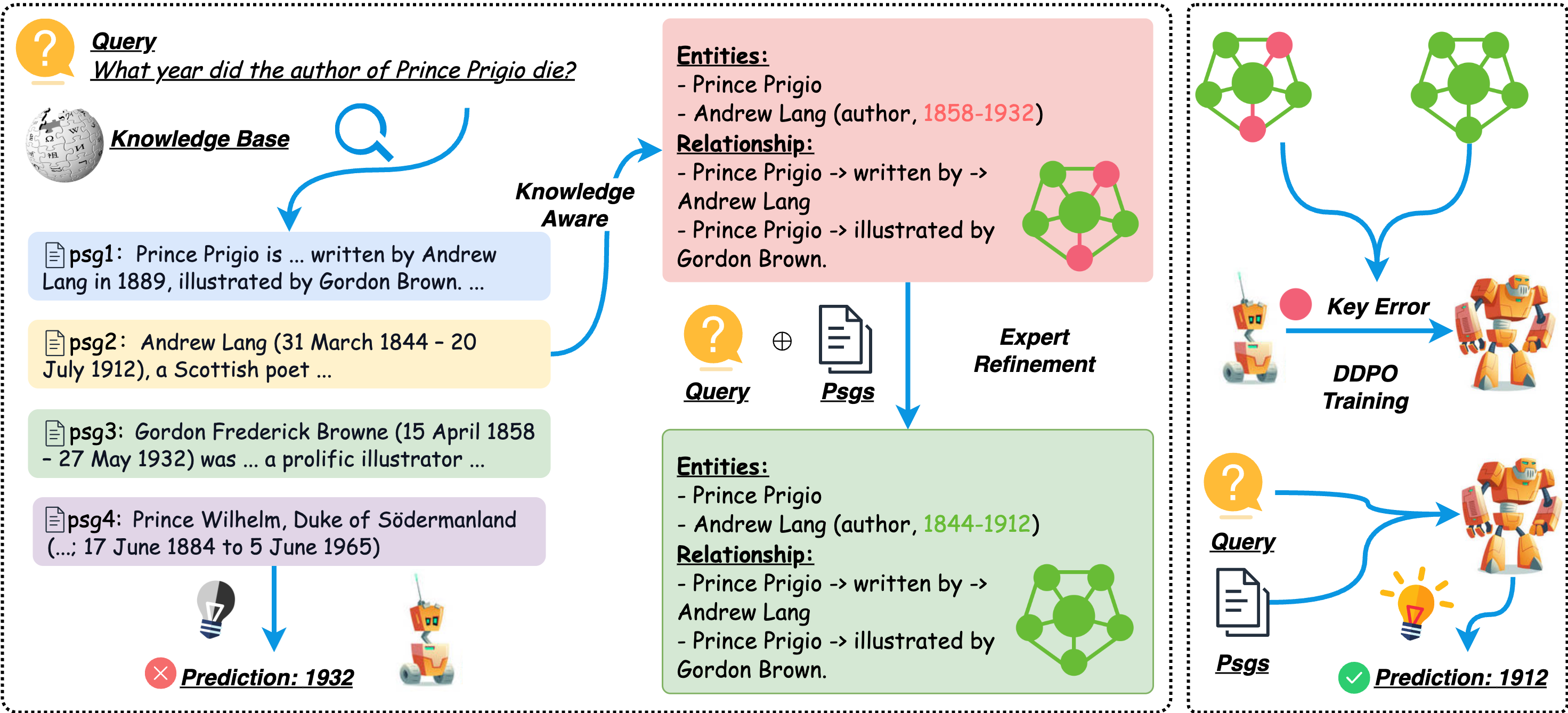}
    \caption{Overview of KARE-RAG. The left panel illustrates the data generation procedure. Retrieved passages can include sufficient evidence for the query, while the generator may still derive an incorrect response. KARE-RAG prompts the generator to construct a scaffold that localizes the evidence-use error. An expert model then revises the scaffold to form a DDPO preference pair. The right panel shows that the optimized generator returns to standard Vanilla RAG inference after training.}
    \label{fig:kare-rag}
\end{figure*}

To address these issues, we develop KARE-RAG (Knowledge-Aware Refinement and Enhancement for RAG), a scaffolded training strategy that teaches models to better utilize retrieved documents. We use KARE as a compact name in tables and experimental discussions. Figure~\ref{fig:kare-rag} provides a schematic overview of the framework. Rather than supervising only final answers, KARE-RAG introduces temporary training scaffolds that make evidence organization explicit during data construction. These scaffolds act as editable intermediate workspaces. An advanced LLM refines localized factual errors while preserving the surrounding organization, yielding high-quality preference pairs that differ mainly in critical regions. For optimization, we adopt Dense Direct Preference Optimization (DDPO)~\cite{yu2024rlhf} with an SFT regularizer~\cite{liu2024provably} and assign token-level weights to edited scaffold regions. Crucially, the scaffolds are used only for alignment. During inference, the optimized model operates through a standard Vanilla RAG pipeline without additional computational overhead.

Empirically, KARE improves transfer across the evaluated QA and relation extraction datasets while retaining the standard Vanilla RAG inference pipeline. It obtains the best results on most cross-dataset metrics, showing that scaffolded alignment can improve evidence use across task formats. Its scaffold-based supervision can also be combined with existing RAG training objectives as a complementary alignment stage.

\section{Related Work} \label{sec:relatedwork}

Retrieval-Augmented Generation (RAG) improves knowledge-intensive generation by conditioning language models on external evidence rather than relying only on parametric memory \cite{lewis2020retrieval,karpukhin2020dense,guu2020retrieval}. Its effectiveness depends not only on retrieving relevant passages but also on whether the generator can use imperfect contexts. Retrieved evidence may be incomplete, irrelevant, conflicting, or difficult to use when useful information is dispersed across a long context \cite{yoran2023making,longpre2022entitybasedknowledgeconflictsquestion,liu2024lost}. Prior work addresses these issues through inference-time mechanisms such as context compression, note construction, and iterative retrieval with reasoning \cite{xu2023recomp,yu-etal-2024-chain,trivedi2022interleaving,xu2024search}. These methods improve how evidence is selected or presented during generation and show why evidence organization matters when retrieval is imperfect.

Another line of work adapts RAG systems through supervised or preference-based training. RA-DiT uses dual instruction tuning to adapt both the language model and the retriever for retrieval-augmented language modeling \cite{lin2023ra}. RAFT fine-tunes models for domain-specific open-book QA with relevant and distractor documents \cite{zhang2024raft}. Self-RAG trains a model to retrieve, generate, and critique using self-reflection signals \cite{asai2023self}. DDR introduces differentiable data rewards to optimize RAG modules according to their effects on system-level performance \cite{li2024rag}, and RPO incorporates retrieval relevance into preference optimization for robust RAG generation \cite{yan-etal-2025-rpo}. These approaches provide strong ways to adapt RAG behavior, but their supervision is mainly tied to final answers, retrieval behavior, or task-level rewards.

Preference optimization provides a natural tool for learning from paired supervision. DPO directly optimizes a language model from preference pairs without training a separate reward model \cite{rafailov2024direct}. Recent work further shows that finer-grained preference signals can be useful when correctness depends on localized errors. RLHF-V collects segment-level correction feedback for hallucination alignment and optimizes with dense direct preference signals \cite{yu2024rlhf}, while Step-DPO applies step-wise preferences to long-chain reasoning rather than judging only complete answers \cite{lai2024step}. This observation is especially relevant for RAG, where a small number of entities, attributes, or relations can determine whether an answer is grounded. It suggests the value of training signals that identify local evidence-use errors rather than judging the full output alone.

\section{Method} \label{sec:method}

In this section, we formalize Knowledge-Aware Refinement and Enhancement for RAG (KARE-RAG), shown in Figure~\ref{fig:kare-rag}. During training data construction, Knowledge-Aware Sampling asks the generator to build a scaffold from the query and retrieved documents before answer generation. The scaffold records how the generator selects and connects evidence, making it possible to locate errors in the intermediate organization. An expert-guided refinement pipeline then edits flawed scaffolds while preserving their overall format, producing preference pairs with localized factual differences. We train the generator with Dense Direct Preference Optimization (DDPO), where token-level weights are assigned to the edited regions. After training, the scaffold interface is removed. The model receives the same inputs as standard Vanilla RAG and generates answers directly. The following sections detail each component.

\subsection{Knowledge-Aware Sampling}\label{sec:pipeline}
In a standard Retrieval-Augmented Generation (RAG) process, the LLM receives a query $q$ and a set of retrieved documents $D=\{d_1,\dots,d_n\}$ to generate an answer. Retrieved documents may contain irrelevant, incomplete, or conflicting information, which can affect generation quality.

During training, Knowledge-Aware Sampling asks the generator to organize the query and retrieved documents into a temporary scaffold before producing the final answer. The scaffold records which evidence the model selects and how the selected evidence is connected. In our main implementation, it takes the form of a lightweight graph-style sketch. We denote this scaffold construction step as
\begin{equation}
    s = \operatorname{LLM}(\mathrm{Instruct}_{\mathrm{scaf}}, q \oplus D)
\end{equation}
where $\oplus$ denotes concatenation, $\mathrm{Instruct}_{\mathrm{scaf}}$ is the scaffold construction instruction, and $s$ denotes the resulting scaffold.

This scaffold provides the refinement stage with a concrete object to inspect. Compared with final-answer supervision, it exposes intermediate evidence organization and separates factual mistakes from unrelated wording differences. When the scaffold uses a structured format, its discrete units also make it easier to identify which entities, attributes, or relations require correction. This property is important for constructing preference pairs that differ in small factual regions rather than in broad surface form.

\subsection{Scaffold Refinement}\label{sec:data_gen}

Large language models can often produce format-valid scaffolds, but these scaffolds may still contain localized factual inaccuracies when processing retrieved documents. Scaffold Refinement constructs preference pairs by correcting such errors while preserving the surrounding scaffold structure. For each query, we generate an initial scaffold $s^-$ and use it to produce an answer, retaining only cases where the answer is incorrect. We then verify whether the retrieved documents support the gold answer. If the evidence is sufficient, an expert model minimally edits $s^-$ into $s^+$ under document-grounded constraints while preserving the scaffold format. The pair $(s^+,s^-)$ is retained only when $s^+$ leads the generator to the correct answer. Appendix~\ref{app:refinement_algorithm} provides the full procedure.

\subsection{Optimization Approach}

The scaffold pairs produced by the refinement stage usually differ only in small factual regions. Standard DPO \cite{rafailov2024direct} treats all output tokens uniformly, so the signal from a corrected entity, attribute, or relation can be diluted by many unchanged scaffold tokens. We therefore adopt Dense Direct Preference Optimization (DDPO) \cite{yu2024rlhf}, which keeps the pairwise preference objective but computes scaffold scores with higher weights on edited tokens. Let $x$ denote the training input and $(s^+,s^-)$ denote the refined and flawed scaffold pair. The weighted DPO objective is
\begin{equation}
\mathcal{L}_{\mathrm{DDPO}}
= -\mathbb{E}_{(x,s^+,s^-)}
\left[\log\sigma\left(\beta \Delta^w\right)\right],
\end{equation}
where
\begin{equation}
\begin{aligned}
\Delta^w(x,s^+,s^-)
&= \ell_\theta^w(s^+ \mid x)-\ell_\theta^w(s^- \mid x) \\
&\quad - \ell_{\mathrm{ref}}^w(s^+ \mid x)+\ell_{\mathrm{ref}}^w(s^- \mid x).
\end{aligned}
\end{equation}

We compute the scaffold score with token-level weights:
\begin{equation}
\begin{aligned}
\ell_\theta^w(s \mid x)
&= \sum_{t=1}^{T} w_t \log p_\theta(s_t \mid x,s_{<t}),\\
w_t &= 1 + (\gamma-1)\mathbf{1}[t \in C].
\end{aligned}
\end{equation}
Here $C$ is the set of modified token positions, and $\gamma>1$ controls how strongly these tokens are emphasized. We obtain $C$ by applying a sequence-level diff between $s^-$ and $s^+$, and then mapping the edited spans to token indices. This affects only training loss computation and introduces no inference overhead.

Because preference optimization can suffer from overoptimization \cite{liu2024provably}, we add an SFT regularizer on the refined scaffold to prevent reward collapse. The final loss is
\begin{equation}
    \mathcal{L} = \mathcal{L}_{\mathrm{DDPO}} - \alpha\sum_{t=1}^{T} \log p_\theta(s_t^+ \mid x,s_{<t}^+),
\end{equation}
where $\alpha$ controls the strength of the regularization. This term anchors the policy to valid document-grounded scaffolds while DDPO focuses the preference signal on edited regions.

Although the optimization target is defined over scaffold outputs, KARE does not use scaffolds as a test-time interface. Scaffold generation serves as an auxiliary training target that exposes localized evidence-use corrections during alignment. These corrections update the same generator parameters used for standard RAG answering. Once alignment is complete, we remove the scaffold instruction and query the model with the standard RAG prompt. The model therefore receives the same input format as Vanilla RAG and directly generates an answer. The pipeline-transfer results in Appendix~\ref{app:pipeline_generalization} support that the learned behavior is not tied to a specific scaffold prompt.

\section{Experimental Settings} \label{sec:experiments}

\begin{table*}[]
\centering
\resizebox{\textwidth}{!}{
\begin{tabular}{lcccccccccccc}
\toprule
\multicolumn{1}{c}{\multirow{3}{*}{\textbf{Method}}} & \multicolumn{2}{c}{\textbf{In Domain}} & \multicolumn{10}{c}{\textbf{Out Of Domain}} \\ 
\cmidrule(lr){2-3} \cmidrule(lr){4-13} 
\multicolumn{1}{c}{} & \multicolumn{2}{c}{\textbf{Musique}} & \multicolumn{2}{c}{\textbf{NQ}} & \multicolumn{2}{c}{\textbf{HotpotQA}} & \multicolumn{2}{c}{\textbf{PopQA}} & \multicolumn{2}{c}{\textbf{TruthfulQA}} & \multicolumn{2}{c}{\textbf{Zero-shot RE}} \\
\cmidrule(lr){2-3} \cmidrule(lr){4-5} \cmidrule(lr){6-7} \cmidrule(lr){8-9} \cmidrule(lr){10-11} \cmidrule(lr){12-13}
\multicolumn{1}{c}{} & \textbf{EM} & \multicolumn{1}{c}{\textbf{F1}} & \textbf{EM} & \textbf{F1} & \textbf{EM} & \textbf{F1} & \textbf{EM} & \textbf{F1} & \textbf{BLEU} & \textbf{Rouge-1} & \textbf{F1} & \textbf{Precision} \\ 
\midrule
Vanilla RAG & 6.00 & 12.37 & 34.64 & 48.30 & 29.52 & 40.66 & 35.43 & 44.10 & 5.43 & 15.03 & 51.62 & 49.30 \\
Realm & 6.08 & 13.46 & 29.02 & 39.21 & 24.94 & 35.18 & 31.20 & 36.73 & 2.63 & 10.39 & 40.23 & 39.52 \\
RetRobust & 5.25 & 12.62 & 28.99 & 39.06 & 24.27 & 34.08 & 32.52 & 36.75 & 3.53 & 10.96 & 37.97 & 36.90 \\
RAFT & 6.04 & 15.07 & 27.92 & 41.63 & 24.96 & 38.71 & 35.82 & 44.39 & \textbf{9.25} & \textbf{24.49} & \underline{57.95} & \underline{54.96} \\
RPO & 6.37 & 13.34 & 28.03 & 40.84 & 28.25 & 40.21 & 32.87 & 42.69 & 4.15 & 15.27 & 53.50 & 49.52 \\
RA-DiT & \textbf{8.11} & \underline{17.83} & 33.08 & 45.66 & 27.17 & 39.33 & 36.36 & 43.78 & 5.69 & 14.70 & 45.16 & 44.04 \\
DDR & 7.82 & \textbf{19.18} & \underline{34.78} & \underline{48.85} & \underline{30.58} & \underline{42.59} & \underline{40.63} & \underline{45.90} & 1.68 & 8.41 & 55.34 & 49.12 \\
KARE & \underline{8.02} & 15.75 & \textbf{37.86} & \textbf{50.84} & \textbf{32.36} & \textbf{44.29} & \textbf{40.88} & \textbf{47.77} & \underline{7.42} & \underline{17.76} & \textbf{59.52} & \textbf{58.00} \\ 
\bottomrule
\end{tabular}
}
\caption{Overall performance on Llama-3.1-8B-Instruct. The \textbf{best} result in each column is highlighted in bold, and the \underline{second-best} result is underlined.}
\label{tab:all}
\end{table*}

In this section we first introduce our experimental settings, including datasets, evaluation metrics, and implementation details.

\textbf{Dataset.} In our experiments, we constructed training data from Musique\cite{trivedi2022musiquemultihopquestionssinglehop}, a challenging multi-hop QA dataset that requires models to identify and connect multiple pieces of evidence. This makes it well suited for constructing and refining training-time scaffolds that capture evidence organization. Using FlashRAG\cite{FlashRAG}, we used Wikipedia as the document source and bge-large-en-v1.5\cite{bge_embedding} as the retrieval engine.

Using the process described in Section~\ref{sec:data_gen}, we generated 2,401 training pairs from the 19,938 Musique entries. Appendix~\ref{app:scaffold_data_quality} reports the full filtering funnel and the reasons for discarding examples. Musique is used for the main training pipeline, and its development set serves as the in-domain test set. To examine whether data construction depends on this source dataset, we also build KARE training data from Natural Questions and HotpotQA, with results reported in Appendix~\ref{app:dataset_selection}. To evaluate out-of-distribution (OOD) performance, we incorporate Natural Questions\cite{kwiatkowski2019natural} and PopQA\cite{mallen2022not} for single-hop retrieval, HotpotQA\cite{yang2018hotpotqa} for multi-hop reasoning, TruthfulQA\cite{lin2021truthfulqa} for truthfulness-oriented QA under generation-style answer matching, and Zero-shot RE\cite{levy2017zero} for slot-filling. We also include domain-shift tests on RAGBench\cite{friel2024ragbench} domains covering legal, finance, and biomedical QA, with details reported in Appendix~\ref{app:domain_shift}. This diverse setup supports a systematic analysis of generalization across task formats and domain settings.

\textbf{Evaluation.} Following FlashRAG\cite{FlashRAG}, we employ exact match (EM) and F1 score as evaluation metrics for question answering. For TruthfulQA generation evaluation, we use BLEU and ROUGE-1 scores to measure answer matching. For zero-shot relation extraction, we use F1 score and precision.

\textbf{Baselines.} We compare against baselines that cover the main alternatives to scaffolded alignment, without imposing a strict taxonomy. Vanilla RAG directly conditions the backbone model on retrieved documents without parameter updates. RA-DiT\cite{lin2023ra} and RAFT\cite{zhang2024raft} represent retrieval-conditioned supervised adaptation. RA-DiT instruction-tunes the model to use retrieved information, while RAFT fine-tunes on question-document-answer examples with relevant and distractor contexts for domain-specific RAG. DDR\cite{li2024rag} and RPO\cite{yan-etal-2025-rpo} represent preference or reward-based RAG optimization. DDR uses final-task rewards to align RAG modules, and RPO optimizes preferences tied to retrieval relevance and knowledge conflicts. We also include retrieval-context construction baselines. The REALM-style oracle-context variant\cite{guu2020retrieval} trains with answer-supporting evidence, while RetRobust\cite{yoran2023making} trains with mixtures of relevant and irrelevant contexts. Since several original methods update retrievers or assume different RAG modules, all baselines are adapted to our fixed-retriever, generator-only setting using the reconstructed dataset. The comparison concerns generator adaptation under shared inputs and training resources, not exact reproduction of each published system. Details are provided in Appendix~\ref{app:additional_baseline_settings}.

\textbf{Implementation Details.} We use Llama-3.1-8B-Instruct\cite{grattafiori2024llama} as the main backbone. We also evaluate Llama-3.2-3B-Instruct and Qwen2.5-14B-Instruct\cite{Yang2024Qwen25TR}, with results reported in Appendix~\ref{app:main_other_backbones}. All methods retrieve five documents per query using the fixed retriever described above. Training is implemented with TRL\cite{vonwerra2022trl} and LoRA\cite{hu2021lora}. We train for one epoch with a learning rate of 5e-5. For preference optimization, we set $\beta=0.1$ and use $\alpha=0.1$ and $\gamma=1.1$ in DDPO. All KARE scores are averaged over three runs. We use GPT-4o-mini as the default expert model for scaffold refinement. Scaffold construction is a one-time preprocessing step. With GPT-4o-mini, generating 2,401 valid pairs costs about \$20 and takes 8--10 hours. Code and prompts for data construction, training, and evaluation will be released at \url{https://github.com/thunlp/KARE-RAG}.

\section{Evaluation Results} \label{sec:evaluateresult}

We compare KARE with the baseline models and then examine the effects of the scaffold and optimization design.
\begin{table*}[]
\centering
\resizebox{\textwidth}{!}{
\begin{tabular}{lcccccccccccc}
\toprule
\multicolumn{1}{c}{\multirow{3}{*}{\textbf{Format}}} & \multicolumn{2}{c}{\textbf{In Domain}} & \multicolumn{10}{c}{\textbf{Out Of Domain}} \\ 
\cmidrule(lr){2-3} \cmidrule(lr){4-13}
\multicolumn{1}{c}{} & \multicolumn{2}{c}{\textbf{Musique}} & \multicolumn{2}{c}{\textbf{NQ}} & \multicolumn{2}{c}{\textbf{HotpotQA}} & \multicolumn{2}{c}{\textbf{PopQA}} & \multicolumn{2}{c}{\textbf{TruthfulQA}} & \multicolumn{2}{c}{\textbf{Zero-shot RE}} \\
\cmidrule(lr){2-3} \cmidrule(lr){4-5} \cmidrule(lr){6-7} \cmidrule(lr){8-9} \cmidrule(lr){10-11} \cmidrule(lr){12-13}
\multicolumn{1}{c}{} & \textbf{EM} & \textbf{F1} & \textbf{EM} & \textbf{F1} & \textbf{EM} & \textbf{F1} & \textbf{EM} & \textbf{F1} & \textbf{BLEU} & \textbf{Rouge-1} & \textbf{F1} & \textbf{Precision} \\ 
\midrule
None (Vanilla RAG) & 6.00 & 12.37 & 34.64 & 48.30 & 29.52 & 40.66 & 35.43 & 44.10 & 5.43 & 15.03 & 51.62 & 49.30 \\
Summary & 5.67 & 14.20 & 32.77 & 47.11 & 24.43 & 37.44 & 39.22 & 46.72 & \textbf{7.75} & \textbf{21.81} & 51.71 & 47.47 \\
Keypoints & \underline{7.36} & \textbf{16.31} & \underline{34.73} & \underline{49.06} & \underline{30.07} & \underline{42.93} & \textbf{41.24} & \textbf{48.38} & 6.75 & \underline{19.46} & \underline{55.01} & \underline{51.74} \\
Graph (Ours) & \textbf{8.02} & \underline{15.75} & \textbf{37.86} & \textbf{50.84} & \textbf{32.36} & \textbf{44.29} & \underline{40.88} & \underline{47.77} & \underline{7.42} & 17.76 & \textbf{59.52} & \textbf{58.00} \\ 
\bottomrule
\end{tabular}
}
\caption{Ablation study on different training scaffold formats. We compare the impact of structuring retrieved information as Summaries, Keypoints, or Graphs (Ours) against the Vanilla RAG baseline. The \textbf{best} result in each column is highlighted in bold, and the \underline{second-best} result is underlined.}
\label{tab:ablation_format}
\end{table*}

\subsection{Main Results}

Table~\ref{tab:all} shows that KARE obtains the best results on NQ, HotpotQA, PopQA, and zero-shot relation extraction. These datasets cover single-hop QA, multi-hop QA, and relation extraction, so the gains are not confined to one task format. On TruthfulQA, KARE obtains the second-best BLEU and ROUGE-1 scores after RAFT.

Musique, the in-domain dataset, shows a different pattern. RA-DiT obtains the best EM and DDR the best F1, while KARE performs better on most cross-dataset metrics. The results indicate a trade-off between fitting the training distribution and transferring to other datasets.

Appendix~\ref{app:matched_supervision} reports controls that keep the training instances, expert access, filtering, backbone, and optimization settings fixed while replacing scaffold supervision with answer-level or free-form rationale targets. KARE performs better than these controls on all reported metrics.

Section~\ref{sec:hybrid_training} examines whether scaffolded alignment can be combined with another RAG training objective.

\subsection{Scaffold Format Matters}

To investigate how scaffold representation affects training, we compare three formats with different degrees of local structure. Summary consolidates retrieved information into a paragraph. Keypoints list salient evidence units line by line. Graphs represent entities and relations in a more explicit structure. All experiments use Llama-3.1-8B-Instruct, and the prompts are detailed in Appendix~\ref{app:prompts}.

Table~\ref{tab:ablation_format} shows that KARE is not tied to a single scaffold format. Keypoints give the best PopQA scores, while graph-structured scaffolds give the best results on NQ, HotpotQA, and zero-shot RE and remain competitive elsewhere. We use the graph-structured scaffold as the default because it performs well on tasks that require multi-hop evidence or fine-grained relations. The framework itself does not require a graph scaffold.

\begin{table*}[t]
\centering
\resizebox{\textwidth}{!}{
\begin{tabular}{lcccccccccccc}
\toprule
\multicolumn{1}{c}{\multirow{3}{*}{\textbf{Optimization Method}}} & \multicolumn{2}{c}{\textbf{In Domain}} & \multicolumn{10}{c}{\textbf{Out Of Domain}} \\
\cmidrule(lr){2-3} \cmidrule(lr){4-13}
\multicolumn{1}{c}{} & \multicolumn{2}{c}{\textbf{Musique}} & \multicolumn{2}{c}{\textbf{NQ}} & \multicolumn{2}{c}{\textbf{HotpotQA}} & \multicolumn{2}{c}{\textbf{PopQA}} & \multicolumn{2}{c}{\textbf{TruthfulQA}} & \multicolumn{2}{c}{\textbf{Zero-shot RE}} \\
\cmidrule(lr){2-3} \cmidrule(lr){4-5} \cmidrule(lr){6-7} \cmidrule(lr){8-9} \cmidrule(lr){10-11} \cmidrule(lr){12-13}
\multicolumn{1}{c}{} & \textbf{EM} & \textbf{F1} & \textbf{EM} & \textbf{F1} & \textbf{EM} & \textbf{F1} & \textbf{EM} & \textbf{F1} & \textbf{BLEU} & \textbf{Rouge-1} & \textbf{F1} & \textbf{Precision} \\
\midrule
None (Vanilla RAG) & 6.00 & 12.37 & 34.64 & 48.30 & 29.52 & 40.66 & 35.43 & 44.10 & 5.43 & 15.03 & 51.62 & 49.30 \\
SFT & 4.88 & 10.46 & 33.62 & 44.10 & 23.69 & 32.77 & 35.88 & 40.73 & 3.12 & 9.63 & 47.50 & 45.05 \\
KARE (DDPO) & \textbf{8.02} & \textbf{15.75} & \textbf{37.86} & \textbf{50.84} & \textbf{32.36} & \textbf{44.29} & \textbf{40.88} & \textbf{47.77} & 7.42 & 17.76 & \textbf{59.52} & \textbf{58.00} \\
\quad - w/o SFT Loss & 6.54 & 13.92 & 33.08 & 46.74 & 27.51 & 40.43 & 37.18 & 44.34 & 6.34 & \textbf{18.54} & 54.36 & 51.92 \\
\quad - w/o Token Weight & 7.73 & 15.75 & 37.40 & 50.57 & 32.26 & 43.83 & 40.27 & 47.39 & \textbf{7.64} & 17.00 & 58.72 & 57.10 \\
\bottomrule
\end{tabular}
}
\caption{Ablation study on optimization approaches, including KARE (DDPO), SFT, and KARE variants without the SFT loss or token weighting. ``None (Vanilla RAG)'' denotes the baseline without training.}
\label{tab:ablation-ddpo}
\end{table*}

\begin{table*}[]
\centering
\resizebox{\textwidth}{!}{
\begin{tabular}{lcccccccccccc}
\toprule
\multicolumn{1}{c}{\multirow{3}{*}{\textbf{Method}}} & \multicolumn{2}{c}{\textbf{In Domain}} & \multicolumn{10}{c}{\textbf{Out Of Domain}} \\ 
\cmidrule(lr){2-3} \cmidrule(lr){4-13}
\multicolumn{1}{c}{} & \multicolumn{2}{c}{\textbf{Musique}} & \multicolumn{2}{c}{\textbf{NQ}} & \multicolumn{2}{c}{\textbf{HotpotQA}} & \multicolumn{2}{c}{\textbf{PopQA}} & \multicolumn{2}{c}{\textbf{TruthfulQA}} & \multicolumn{2}{c}{\textbf{Zero-shot RE}} \\
\cmidrule(lr){2-3} \cmidrule(lr){4-5} \cmidrule(lr){6-7} \cmidrule(lr){8-9} \cmidrule(lr){10-11} \cmidrule(lr){12-13}
\multicolumn{1}{c}{} & \textbf{EM} & \textbf{F1} & \textbf{EM} & \textbf{F1} & \textbf{EM} & \textbf{F1} & \textbf{EM} & \textbf{F1} & \textbf{BLEU} & \textbf{Rouge-1} & \textbf{F1} & \textbf{Precision} \\ 
\midrule
Vanilla RAG & 6.00 & 12.37 & 34.64 & 48.30 & 29.52 & 40.66 & 35.43 & 44.10 & 5.43 & 15.03 & 51.62 & 49.30 \\
DDR & 7.82 & \underline{19.18} & 34.78 & 48.85 & 30.58 & 42.59 & \underline{40.63} & 45.90 & 1.68 & 8.41 & 55.34 & 49.12 \\
KARE & \underline{8.02} & 15.75 & \underline{37.86} & \underline{50.84} & \underline{32.36} & \underline{44.29} & \textbf{40.88} & \underline{47.77} & \textbf{7.42} & \textbf{17.76} & \underline{59.52} & \underline{58.00} \\
DDR + KARE & \textbf{9.01} & \textbf{20.88} & \textbf{38.49} & \textbf{51.48} & \textbf{33.06} & \textbf{45.13} & 40.53 & \textbf{48.30} & 2.30 & 10.85 & \textbf{60.67} & \textbf{59.74} \\ 
\bottomrule
\end{tabular}
}
\caption{Performance comparison of the hybrid training approach (DDR + KARE) against baselines on Llama-3.1-8B-Instruct. "DDR + KARE" denotes the model initialized with DDR weights and further fine-tuned using the KARE method. The \textbf{best} result in each column is highlighted in bold, and the \underline{second-best} result is underlined.}
\label{tab:hybrid}
\end{table*}

\subsection{Optimization Ablation}

Table~\ref{tab:ablation-ddpo} isolates two parts of the optimization objective: token-level weighting in DDPO and the auxiliary SFT regularizer. It reports the same point estimates as the main experiments. Appendix~\ref{app:optimization_variance} gives the mean and standard deviation over three independent runs for the KARE variants.

The SFT baseline trains on the refined scaffolds from the same small, filtered set rather than on final answers. All variants use the direct-answer RAG prompt at test time. Its lower scores may arise from the lack of rejected-scaffold supervision, the mismatch between scaffold training and direct-answer inference, or both. This comparison does not show that SFT is generally harmful.

Removing token weighting slightly lowers most metrics, although Musique F1 is unchanged and TruthfulQA BLEU improves. Chosen and rejected scaffolds often differ in only a few factual spans, so a uniform preference loss also assigns weight to the many tokens shared by the pair. DDPO instead gives more weight to the edited spans. The three-run results in Appendix~\ref{app:optimization_variance} show the same broad pattern, with lower variation for the full objective on most metrics.

Removing the SFT term lowers most scores and increases run-to-run variation. The training curves in Appendix~\ref{app:training_curves} show that the DPO margin grows while the implicit rewards for both chosen and rejected responses fall, which is consistent with policy drift. The SFT term keeps the policy closer to the refined scaffold distribution and reduces this drift.

\subsection{Exploring Hybrid Training}
\label{sec:hybrid_training}

KARE can be combined with other RAG training objectives. We test this by starting from a DDR-trained model and applying KARE in a second LoRA fine-tuning stage without changing inference. Table~\ref{tab:hybrid} reports the resulting performance. This experiment shows that the two training objectives can be used together.

\begin{table*}[]
\centering
\resizebox{\textwidth}{!}{
\begin{tabular}{lcccccccccccc}
\toprule
\multicolumn{1}{c}{\multirow{3}{*}{\textbf{Refinement Expert}}} & \multicolumn{2}{c}{\textbf{In Domain}} & \multicolumn{10}{c}{\textbf{Out Of Domain}} \\ 
\cmidrule(lr){2-3} \cmidrule(lr){4-13}
\multicolumn{1}{c}{} & \multicolumn{2}{c}{\textbf{Musique}} & \multicolumn{2}{c}{\textbf{NQ}} & \multicolumn{2}{c}{\textbf{HotpotQA}} & \multicolumn{2}{c}{\textbf{PopQA}} & \multicolumn{2}{c}{\textbf{TruthfulQA}} & \multicolumn{2}{c}{\textbf{Zero-shot RE}} \\
\cmidrule(lr){2-3} \cmidrule(lr){4-5} \cmidrule(lr){6-7} \cmidrule(lr){8-9} \cmidrule(lr){10-11} \cmidrule(lr){12-13}
\multicolumn{1}{c}{} & \textbf{EM} & \textbf{F1} & \textbf{EM} & \textbf{F1} & \textbf{EM} & \textbf{F1} & \textbf{EM} & \textbf{F1} & \textbf{BLEU} & \textbf{Rouge-1} & \textbf{F1} & \textbf{Precision} \\ 
\midrule
None (Vanilla RAG) & 6.00 & 12.37 & 34.64 & 48.30 & 29.52 & 40.66 & 35.43 & 44.10 & 5.43 & 15.03 & 51.62 & 49.30 \\
Qwen2.5-14B-Instruct & 6.95 & 15.16 & 36.35 & 49.17 & 31.48 & 43.41 & 38.61 & 45.92 & 7.35 & \textbf{18.05} & 55.16 & 53.07 \\
GPT-4o-mini (sampled) & 7.36 & 15.23 & 36.13 & 49.21 & 31.49 & 43.58 & 38.28 & 46.13 & 7.27 & 17.45 & 56.12 & 54.16 \\
GPT-4o-mini & \textbf{8.02} & 15.75 & \textbf{37.86} & \textbf{50.84} & 32.36 & 44.29 & \textbf{40.88} & \textbf{47.77} & 7.42 & 17.76 & 59.52 & 58.00 \\
GPT-4o & 7.82 & \textbf{15.98} & 37.70 & 50.51 & \textbf{32.75} & \textbf{44.36} & 40.37 & 47.44 & \textbf{7.59} & 17.32 & \textbf{59.73} & \textbf{58.42} \\ 
\bottomrule
\end{tabular}
}
\caption{Impact of different expert models on KARE performance. Qwen2.5-14B-Instruct and the sampled GPT-4o-mini setting each use 1,043 accepted pairs; the full GPT-4o-mini setting uses 2,401 pairs. The sampled row is averaged over three training runs. All experiments use Llama-3.1-8B-Instruct as the generator.}
\label{tab:expert_impact}
\end{table*}

\subsection{Expert Model Diagnostics}\label{sec:expert_diagnostics}

We compare Qwen2.5-14B-Instruct, GPT-4o-mini, and GPT-4o as refinement experts in the same data construction pipeline. Qwen2.5-14B-Instruct produces 1,043 valid pairs, while GPT-4o-mini and GPT-4o each produce around 2,400. We also sample 1,043 pairs from the GPT-4o-mini data to control for the number of accepted pairs and average the downstream results over three training runs. Table~\ref{tab:expert_impact} shows similar results for the matched Qwen2.5-14B and GPT-4o-mini sets on most tasks. Training on all GPT-4o-mini pairs gives better overall performance, which suggests that pair count matters more than expert identity in this comparison. GPT-4o and the full GPT-4o-mini setting give similar results, so we use GPT-4o-mini as the default expert because it costs less.

These results also reflect strict filtering in our data pipeline. A pair is retained only when the retrieved documents support the answer and the refined scaffold leads the generator to the correct output. This criterion reduces data volume, but keeps the final supervision concentrated on high-quality instances. Appendix~\ref{app:scaffold_data_quality} reports the complete filtering funnel, failure breakdown, manual audit, and corruption analysis.

\subsection{Generality Across Retrievers, Domains, and Pipelines}

We test the same trained generator under different retrievers, domains, and inference pipelines.

KARE is trained with bge-large-en-v1.5 retrieval. At inference, we replace this retriever with e5-base-v2 or BM25 without retraining the generator. KARE improves over the no-training baseline with both alternative retrievers, although the absolute scores vary with retriever quality. Full results are reported in Appendix~\ref{app:retriever_transfer}.

\begin{table*}[t]
\centering
\resizebox{0.86\textwidth}{!}{
\begin{tabular}{lccccccccc}
\toprule
\multicolumn{1}{c}{\multirow{2}{*}{\textbf{Method}}} & \multicolumn{3}{c}{\textbf{CUAD (Legal)}} & \multicolumn{3}{c}{\textbf{FinQA (Finance)}} & \multicolumn{3}{c}{\textbf{PubMedQA (Biomedical)}} \\
\cmidrule(lr){2-4} \cmidrule(lr){5-7} \cmidrule(lr){8-10}
& \textbf{F1} & \textbf{Adh.} & \textbf{Comp.} & \textbf{F1} & \textbf{Adh.} & \textbf{Comp.} & \textbf{F1} & \textbf{Adh.} & \textbf{Comp.} \\
\midrule
Vanilla RAG & 9.01 & 55.88 & 20.18 & 16.98 & 55.27 & 30.89 & 11.40 & 85.92 & 37.16 \\
DDR & 10.70 & 38.04 & 21.60 & 7.80 & 35.74 & 25.81 & 8.15 & 62.41 & 23.22 \\
RAFT & \textbf{12.19} & 45.49 & 17.98 & 6.75 & 33.13 & 28.14 & 17.86 & 77.96 & 44.80 \\
KARE & 11.56 & \textbf{56.41} & \textbf{22.92} & \textbf{19.62} & \textbf{56.09} & \textbf{32.57} & \textbf{18.26} & \textbf{86.20} & \textbf{46.37} \\
\bottomrule
\end{tabular}
}
\caption{Domain-shift evaluation on three RAGBench domains. Adh. and Comp. denote adherence and completeness judged by GPT-4.1-mini. All methods use the same fixed-retriever, generator-only protocol.}
\label{tab:ragbench_domain_shift}
\end{table*}

We also evaluate domain shift on legal, finance, and biomedical tasks from RAGBench. Table~\ref{tab:ragbench_domain_shift} shows that KARE obtains the best result on all six adherence and completeness metrics and on eight of the nine metrics overall. RAFT has a higher CUAD F1. Appendix~\ref{app:domain_shift} details the setup.

KARE also improves over the no-training baselines under Vanilla RAG, CoT, CoN, StructRAG, and IRCoT. These pipelines modify prompting or reasoning, while KARE modifies the generator through training, so the two can be used together. Full results are reported in Appendix~\ref{app:pipeline_generalization}.

Appendix~\ref{app:inference_comparisons} separately compares direct answering with explicit scaffold generation, LLMLingua-2 context compression, and controlled Self-RAG adaptations. These settings modify the inference procedure and are therefore kept separate from the main generator-adaptation comparison.

\subsection{General Ability Diagnostics}\label{sec:general_ability_diagnostics}

Table~\ref{tab:mmlu-test} checks whether KARE affects general ability. Accuracy changes from 68.00 to 67.90 on MMLU\cite{hendrycks2020measuring} and from 43.97 to 44.60 on MMLU-Pro\cite{wang2024mmlu}. These small changes provide no evidence of a substantial loss of general ability on either benchmark.

\begin{table}
\centering
\resizebox{0.45\textwidth}{!}{
\begin{tabular}{lcc}
\toprule
\textbf{Train Method} & \textbf{MMLU (Acc)} & \textbf{MMLU-Pro (Acc)} \\ 
\midrule
None & \textbf{68.00} & 43.97 \\
KARE & 67.90 & \textbf{44.60} \\
\bottomrule
\end{tabular}
}
\caption{General ability evaluation on MMLU and MMLU-Pro datasets using Llama-3.1-8B-Instruct.}
\label{tab:mmlu-test}
\end{table}

\section{Conclusion} \label{sec:conclusion}

This paper presents KARE-RAG, a structured-supervision framework for RAG. KARE builds localized preference pairs from temporary scaffolds, teaching valid evidence organization under retrieval noise. It adopts DDPO to learn from these localized corrections and improves transfer across the evaluated QA and relation extraction datasets while preserving standard RAG inference without extra runtime cost. Its scaffold-based supervision can also complement existing RAG training objectives as an additional alignment stage. These findings suggest that fine-grained scaffolded supervision is promising for reliable knowledge-intensive generation.

\section{Limitations}

\begin{itemize}
    \item \textbf{Retrieval coverage}. KARE improves how the generator uses retrieved evidence, but it still assumes sufficient retrieval coverage. If the retrieved documents do not contain answer-supporting evidence, the refinement stage cannot recover information that is absent from the context.
    \item \textbf{Scaffold design}. The scaffold design space remains only partially explored. We study summaries, keypoints, and graph-structured sketches, so the conclusions may not cover all possible scaffold formats or task-specific decomposition strategies.
    \item \textbf{Evaluation scope}. Our evaluation mainly focuses on English factual RAG. The results may not directly transfer to multilingual RAG or open-ended dialogue, where evidence ambiguity, user intent, and interaction history can be substantially different.
    \item \textbf{Expert-assisted data construction}. KARE relies on an expert model to refine scaffolds and filter valid pairs. Although this is a one-time preprocessing cost and does not affect inference latency, the quality and volume of supervision can depend on expert capability and the scaffold length limit. Strict filtering also concentrates training on cases that can be corrected from the retrieved evidence.
\end{itemize}

\section*{Acknowledgments}

This work was supported by the National Key Research and Development Program of China under Grant No.~2025ZD0215701, the National Natural Science Foundation of China under Grant Nos.~T2341003 and U2336214, and the Beijing Natural Science Foundation under Grant No.~L257019. We also thank the AI9Stars community for its support.

\bibliography{references-camera-p4}

\appendix
\renewcommand{\floatpagefraction}{0.8}
\renewcommand{\dblfloatpagefraction}{0.8}
\section{Appendix}\label{app:appendix}

\subsection{LLM Employment}\label{app:llm_employment}

In the preparation of this paper, we employed large language models (LLMs) exclusively to assist with language-related aspects of the writing process. Specifically, LLMs were used to refine grammatical structures, facilitate translation of certain text segments, and provide final polishing to improve the clarity and readability of the manuscript. No substantial intellectual, methodological, or analytical input was contributed by LLMs, and all scientific content, results, interpretations, and conclusions remain entirely our own.

\subsection{Scaffold Refinement Algorithm}\label{app:refinement_algorithm}

This appendix expands on the filtering and editing process summarized in Section~\ref{sec:data_gen}. For each query-document set, the base generator first produces a scaffold $s^-$ through Knowledge-Aware Sampling, and the same generator uses this scaffold to answer the query. We retain only examples in which the generated answer is incorrect, because these cases expose a concrete scaffold-level failure that can be corrected. We then ask the expert model to judge whether the retrieved documents $D$ contain enough evidence to support the gold answer $y_{\mathrm{gnd}}$. If the evidence is insufficient, the example is discarded rather than repaired, since any correction would require information outside the retrieved documents.

For evidence-sufficient cases, $\mathrm{LLM_{Exp}}$ receives the query, retrieved documents, gold answer, and flawed scaffold. It is instructed to minimally revise $s^-$ while preserving the scaffold format and most of the surrounding structure. The revision should only add, remove, or modify content supported by the retrieved documents. The refined scaffold $s^+$ is then passed back to $\mathrm{LLM_{Gen}}$ for answer generation. We accept $(s^+,s^-)$ only when $s^+$ leads to the correct answer, which keeps the preference pair grounded in the retrieved evidence and aligned with the factual error that caused the original failure.

\begin{algorithm}[t]
\caption{Scaffold Refinement}
\label{alg:data_construction}
\begin{algorithmic}[1]
\STATE Generate initial scaffold $s^-$ using $\mathrm{LLM_{Gen}}$ with Knowledge-Aware Sampling
\IF{$s^-$ produces incorrect answer $y_{\mathrm{err}} \neq y_{\mathrm{gnd}}$}
    \STATE Check document adequacy with $\mathrm{LLM_{Exp}}(q \oplus D \oplus y_{\mathrm{gnd}})$
    \IF{Document $D$ is sufficient}
        \STATE Revise the scaffold with an advanced model $\mathrm{LLM_{Exp}}$
        \begin{equation*}
        \begin{aligned}
            s^+ \leftarrow \mathrm{LLM_{Exp}}(&\mathrm{Instruct}_{\mathrm{revise}}, \\
            &q \oplus D \oplus y_{\mathrm{gnd}} \oplus s^-)
        \end{aligned}
        \end{equation*}
        \STATE Generate final answer $y_{\mathrm{Gen}}$ using $s^+$ with $\mathrm{LLM_{Gen}}$
        \WHILE{$y_{\mathrm{Gen}} \neq y_{\mathrm{gnd}}$ \& \textit{iter} < max\_iter}
            \STATE Further revise $s^+$ using error analysis
            \STATE Generate $y_{\mathrm{Gen}}$ with $\mathrm{LLM_{Gen}}$
        \ENDWHILE
        \IF{$y_{\mathrm{Gen}}=y_{\mathrm{gnd}}$}
            \STATE Add $(s^+, s^-)$ as a contrastive DDPO training sample pair
        \ELSE
            \STATE Discard the sample
        \ENDIF
    \ENDIF
\ENDIF
\end{algorithmic}
\end{algorithm}

\subsection{Additional Baseline Settings}\label{app:additional_baseline_settings}
The baseline results are controlled adaptations rather than exact reproductions of each complete system. We use official code or prompts when available and otherwise follow the procedure described in the paper. We adapt each method to the same Llama backbone, retrieved documents, top-five retrieval, generator-only optimization, and training budget. The published results are not directly comparable because the original systems may use different backbones, retrievers, corpora, or inference modules.

We construct the RA-DiT\cite{lin2023ra} and DDR\cite{li2024rag} training data from Musique. RA-DiT uses the instruction prompts from the paper and five retrieved documents per query. For DDR, we sample several responses at different temperatures from the same retrieved context. We label a response positive if either F1 or EM exceeds its threshold and negative only if both metrics fall below their thresholds. We discard questions without both labels, then pair the highest-scoring positive response with the lowest-scoring negative response for DPO training.

The REALM-style oracle-context variant and RetRobust share the same backbone and training setup: Llama-3.1-8B-Instruct, the fixed bge-large-en-v1.5 retriever, and five retrieved documents per question. Only the generator is updated. We train each model with TRL\cite{vonwerra2022trl} and LoRA\cite{hu2021lora} for one epoch at a learning rate of $5\times10^{-5}$.

RAFT\cite{zhang2024raft} is trained on MuSiQue examples containing answer-supporting and distractor documents, with targets that include a short rationale followed by the answer. RPO\cite{yan-etal-2025-rpo} forms preference pairs from relevant and irrelevant responses under the same retrieved contexts and uses $\beta=0.1$. For the REALM-style oracle-context baseline\cite{guu2020retrieval}, we follow the official implementation's data-construction procedure to build examples with answer-supporting evidence, then fine-tune the generator under the shared configuration above. RetRobust\cite{yoran2023making} follows its official data-construction procedure by mixing relevant and irrelevant documents, while retaining the same generator-only training setup. These results should be read as controlled adaptations under a common experimental setting, not as exact reproductions of the original complete systems.

\subsection{Matched-Budget Supervision Controls}\label{app:matched_supervision}

We compare KARE with three alternatives using the same 2,401 instances, retrieved contexts, GPT-4o-mini access, gold-answer verification and filtering, backbone, and training configuration. Answer-DPO uses expert-corrected final answers without scaffolds. Rationale-DPO uses expert-generated free-form rationales and answers without a structured scaffold. KARE-Answer-Only follows the KARE refinement pipeline but optimizes only the final-answer pairs derived from the refined and flawed scaffolds. Full KARE optimizes the scaffold preference pairs.

As shown in Table~\ref{tab:matched_supervision}, Full KARE performs better than the three answer-level controls on all reported metrics under these matched conditions.

\begin{table*}[t]
\centering
\resizebox{\textwidth}{!}{
\begin{tabular}{lcccccccccccc}
\toprule
\multicolumn{1}{c}{\multirow{2}{*}{\textbf{Method}}} & \multicolumn{2}{c}{\textbf{Musique}} & \multicolumn{2}{c}{\textbf{NQ}} & \multicolumn{2}{c}{\textbf{HotpotQA}} & \multicolumn{2}{c}{\textbf{PopQA}} & \multicolumn{2}{c}{\textbf{TruthfulQA}} & \multicolumn{2}{c}{\textbf{Zero-shot RE}} \\
\cmidrule(lr){2-3} \cmidrule(lr){4-5} \cmidrule(lr){6-7} \cmidrule(lr){8-9} \cmidrule(lr){10-11} \cmidrule(lr){12-13}
& \textbf{EM} & \textbf{F1} & \textbf{EM} & \textbf{F1} & \textbf{EM} & \textbf{F1} & \textbf{EM} & \textbf{F1} & \textbf{BLEU} & \textbf{R-1} & \textbf{F1} & \textbf{Prec.} \\
\midrule
Vanilla RAG & 6.00 & 12.37 & 34.64 & 48.30 & 29.52 & 40.66 & 35.43 & 44.10 & 5.43 & 15.03 & 51.62 & 49.30 \\
Answer-DPO & 6.79 & 15.24 & 30.33 & 44.84 & 27.59 & 40.39 & 33.97 & 44.12 & 3.13 & 11.42 & 51.13 & 48.18 \\
Rationale-DPO & 7.11 & 15.18 & 35.50 & 48.45 & 29.10 & 40.44 & 37.32 & 45.77 & 3.99 & 9.81 & 53.05 & 52.89 \\
KARE-Answer-Only & 6.74 & 14.00 & 30.30 & 44.41 & 27.08 & 39.93 & 33.83 & 43.77 & 3.15 & 11.46 & 50.68 & 47.60 \\
Full KARE & \textbf{8.02} & \textbf{15.75} & \textbf{37.86} & \textbf{50.84} & \textbf{32.36} & \textbf{44.29} & \textbf{40.88} & \textbf{47.77} & \textbf{7.42} & \textbf{17.76} & \textbf{59.52} & \textbf{58.00} \\
\bottomrule
\end{tabular}
}
\caption{Matched-budget supervision controls. All trained variants use the same instances, retrieved contexts, expert access, verification and filtering, backbone, and training configuration.}
\label{tab:matched_supervision}
\end{table*}

\subsection{Optimization Variance on the Main Backbone}\label{app:optimization_variance}

Table~\ref{tab:ablation-ddpo-std} reports mean and standard deviation over three independent runs for the KARE variants on Llama-3.1-8B-Instruct. These statistics supplement the point estimates in Table~\ref{tab:ablation-ddpo}. Given the small number of runs, we treat the comparison as descriptive and do not make a statistical significance claim.

\begin{table*}[t]
\centering
\resizebox{\textwidth}{!}{
\begin{tabular}{lcccccccccccc}
\toprule
\multicolumn{1}{c}{\multirow{3}{*}{\textbf{Optimization Method}}} & \multicolumn{2}{c}{\textbf{In Domain}} & \multicolumn{10}{c}{\textbf{Out Of Domain}} \\
\cmidrule(lr){2-3} \cmidrule(lr){4-13}
\multicolumn{1}{c}{} & \multicolumn{2}{c}{\textbf{Musique}} & \multicolumn{2}{c}{\textbf{NQ}} & \multicolumn{2}{c}{\textbf{HotpotQA}} & \multicolumn{2}{c}{\textbf{PopQA}} & \multicolumn{2}{c}{\textbf{TruthfulQA}} & \multicolumn{2}{c}{\textbf{Zero-shot RE}} \\
\cmidrule(lr){2-3} \cmidrule(lr){4-5} \cmidrule(lr){6-7} \cmidrule(lr){8-9} \cmidrule(lr){10-11} \cmidrule(lr){12-13}
\multicolumn{1}{c}{} & \textbf{EM} & \textbf{F1} & \textbf{EM} & \textbf{F1} & \textbf{EM} & \textbf{F1} & \textbf{EM} & \textbf{F1} & \textbf{BLEU} & \textbf{Rouge-1} & \textbf{F1} & \textbf{Precision} \\
\midrule
KARE (DDPO) & $8.02{\pm}0.23$ & $15.75{\pm}0.35$ & $37.86{\pm}0.48$ & $50.84{\pm}0.61$ & $32.36{\pm}0.24$ & $44.29{\pm}0.24$ & $40.88{\pm}0.33$ & $47.77{\pm}0.20$ & $7.42{\pm}0.69$ & $17.76{\pm}2.59$ & $59.52{\pm}0.50$ & $58.00{\pm}0.68$ \\
\quad - w/o SFT Loss & $6.54{\pm}3.60$ & $13.92{\pm}4.25$ & $33.08{\pm}12.74$ & $46.74{\pm}12.00$ & $27.51{\pm}10.45$ & $40.43{\pm}9.77$ & $37.18{\pm}10.91$ & $44.34{\pm}9.01$ & $6.34{\pm}0.99$ & $18.54{\pm}2.71$ & $54.36{\pm}12.23$ & $51.92{\pm}14.71$ \\
\quad - w/o Token Weight & $7.73{\pm}0.35$ & $15.75{\pm}0.15$ & $37.40{\pm}1.68$ & $50.57{\pm}1.55$ & $32.26{\pm}1.33$ & $43.83{\pm}0.99$ & $40.27{\pm}0.64$ & $47.39{\pm}0.41$ & $7.64{\pm}0.74$ & $17.00{\pm}3.58$ & $58.72{\pm}1.25$ & $57.10{\pm}1.55$ \\
\bottomrule
\end{tabular}
}
\caption{Optimization variance on Llama-3.1-8B-Instruct. Values are mean $\pm$ standard deviation over three independent runs.}
\label{tab:ablation-ddpo-std}
\end{table*}

\subsection{Supplementary Experiments Across Model Sizes}\label{app:main_other_backbones}\label{app:ddpo}

This section reports supplementary experiments on Llama-3.2-3B-Instruct and Qwen2.5-14B-Instruct. We include both the main-comparison results and the optimization ablations to show whether the observations in the main text remain consistent across model sizes.

\begin{table*}[]
\centering
\resizebox{\textwidth}{!}{
\begin{tabular}{lcccccccccccc}
\toprule
\multicolumn{1}{c}{\multirow{3}{*}{\textbf{Method}}} & \multicolumn{2}{c}{\textbf{In Domain}} & \multicolumn{10}{c}{\textbf{Out Of Domain}} \\ 
\cmidrule(lr){2-3} \cmidrule(lr){4-13} 
\multicolumn{1}{c}{} & \multicolumn{2}{c}{\textbf{Musique}} & \multicolumn{2}{c}{\textbf{NQ}} & \multicolumn{2}{c}{\textbf{HotpotQA}} & \multicolumn{2}{c}{\textbf{PopQA}} & \multicolumn{2}{c}{\textbf{TruthfulQA}} & \multicolumn{2}{c}{\textbf{Zero-shot RE}} \\
\cmidrule(lr){2-3} \cmidrule(lr){4-5} \cmidrule(lr){6-7} \cmidrule(lr){8-9} \cmidrule(lr){10-11} \cmidrule(lr){12-13}
\multicolumn{1}{c}{} & \textbf{EM} & \multicolumn{1}{c}{\textbf{F1}} & \textbf{EM} & \textbf{F1} & \textbf{EM} & \textbf{F1} & \textbf{EM} & \textbf{F1} & \textbf{BLEU} & \textbf{Rouge-1} & \textbf{F1} & \textbf{Precision} \\ 
\midrule
\multicolumn{13}{l}{\textit{Llama-3.2-3B-Instruct}} \\ 
\midrule
Vanilla RAG & 4.47 & 9.89 & \underline{33.20} & \underline{45.59} & \textbf{27.89} & 37.75 & \underline{34.32} & 41.60 & \underline{6.36} & \underline{13.78} & 48.70 & 47.40 \\
Realm & 4.59 & 11.42 & 29.30 & 39.37 & 23.44 & 33.40 & 28.41 & 33.31 & 2.34 & 9.36 & 23.04 & 22.66 \\
RetRobust & 5.42 & 12.32 & 29.21 & 39.53 & 22.90 & 32.73 & 29.18 & 34.28 & 3.79 & 10.16 & 22.37 & 21.55 \\
RA-DiT & 5.46 & 13.16 & 26.25 & 37.08 & 21.74 & 31.74 & 33.01 & 37.76 & 3.06 & 10.24 & 32.05 & 31.82 \\
DDR & \textbf{6.00} & \textbf{16.18} & 28.93 & 45.31 & 22.88 & \underline{38.04} & 33.21 & \underline{42.91} & 2.83 & 10.09 & \textbf{52.55} & \textbf{50.76} \\
KARE & \underline{5.59} & \underline{13.28} & \textbf{33.21} & \textbf{46.56} & \underline{27.62} & \textbf{39.59} & \textbf{38.29} & \textbf{45.52} & \textbf{7.58} & \textbf{17.75} & \underline{51.94} & \underline{49.03} \\ 
\midrule
\multicolumn{13}{l}{\textit{Qwen2.5-14B-Instruct}} \\ 
\midrule
Vanilla RAG & 6.66 & 14.58 & 32.67 & 47.01 & 30.64 & 42.31 & 36.38 & 43.98 & \underline{6.90} & \underline{18.15} & 53.26 & 49.27 \\
Realm & 7.94 & 15.33 & \underline{34.37} & \underline{48.08} & 31.48 & 43.78 & 38.14 & 44.55 & 1.49 & 8.50 & 51.46 & \underline{51.41} \\
RetRobust & 7.41 & 15.10 & 33.36 & 43.62 & 31.02 & 41.58 & 31.83 & 36.12 & 1.88 & 8.81 & 43.08 & 42.67 \\
RA-DiT & \underline{9.06} & \underline{18.47} & 33.36 & 43.62 & 31.02 & 41.58 & 37.32 & 44.91 & 1.59 & 9.63 & \underline{54.36} & \textbf{53.91} \\
DDR & \textbf{9.68} & \textbf{19.84} & 32.62 & 45.85 & \underline{32.05} & \underline{44.65} & \textbf{41.08} & \underline{45.87} & 1.14 & 8.30 & 52.08 & 50.13 \\
KARE & 8.32 & 17.48 & \textbf{36.19} & \textbf{49.96} & \textbf{33.96} & \textbf{46.45} & \underline{38.58} & \textbf{46.09} & \textbf{8.80} & \textbf{20.39} & \textbf{54.47} & 50.55 \\ 
\bottomrule
\end{tabular}
}
\caption{Additional main-comparison results on Llama-3.2-3B-Instruct and Qwen2.5-14B-Instruct. The main text focuses on Llama-3.1-8B-Instruct, while these results provide complementary evidence across model scales. The \textbf{best} result in each model block is highlighted in bold, and the \underline{second-best} result is underlined.}
\label{tab:main_other_backbones}
\end{table*}

\begin{figure*}[t]
    \centering
    \begin{subfigure}[b]{0.33\textwidth}
        \centering
        \includegraphics[width=\textwidth]{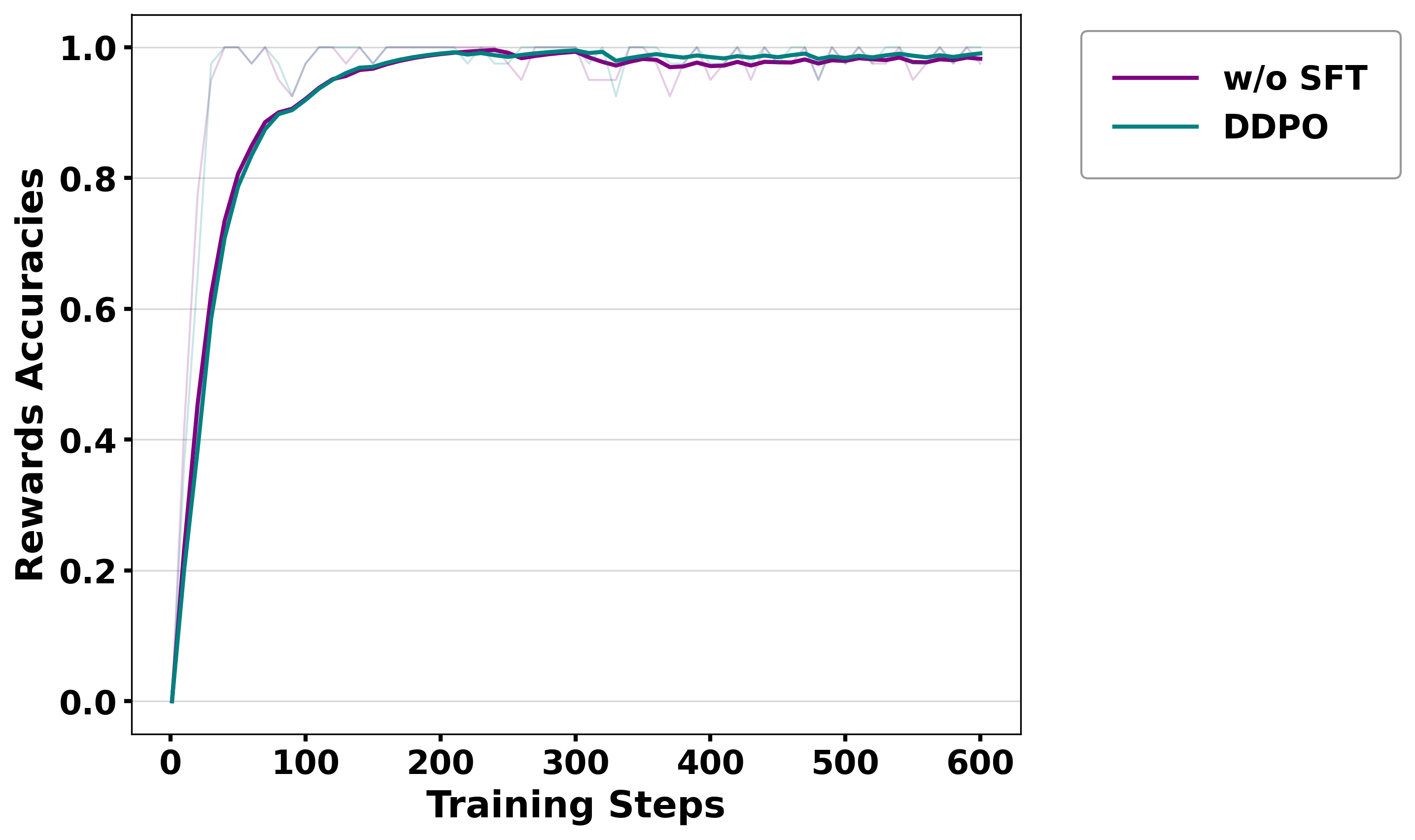}
        \caption{Reward/accuracy}
        \label{fig:reward-acc}
    \end{subfigure}%
    \begin{subfigure}[b]{0.33\textwidth}
        \centering
        \includegraphics[width=\textwidth]{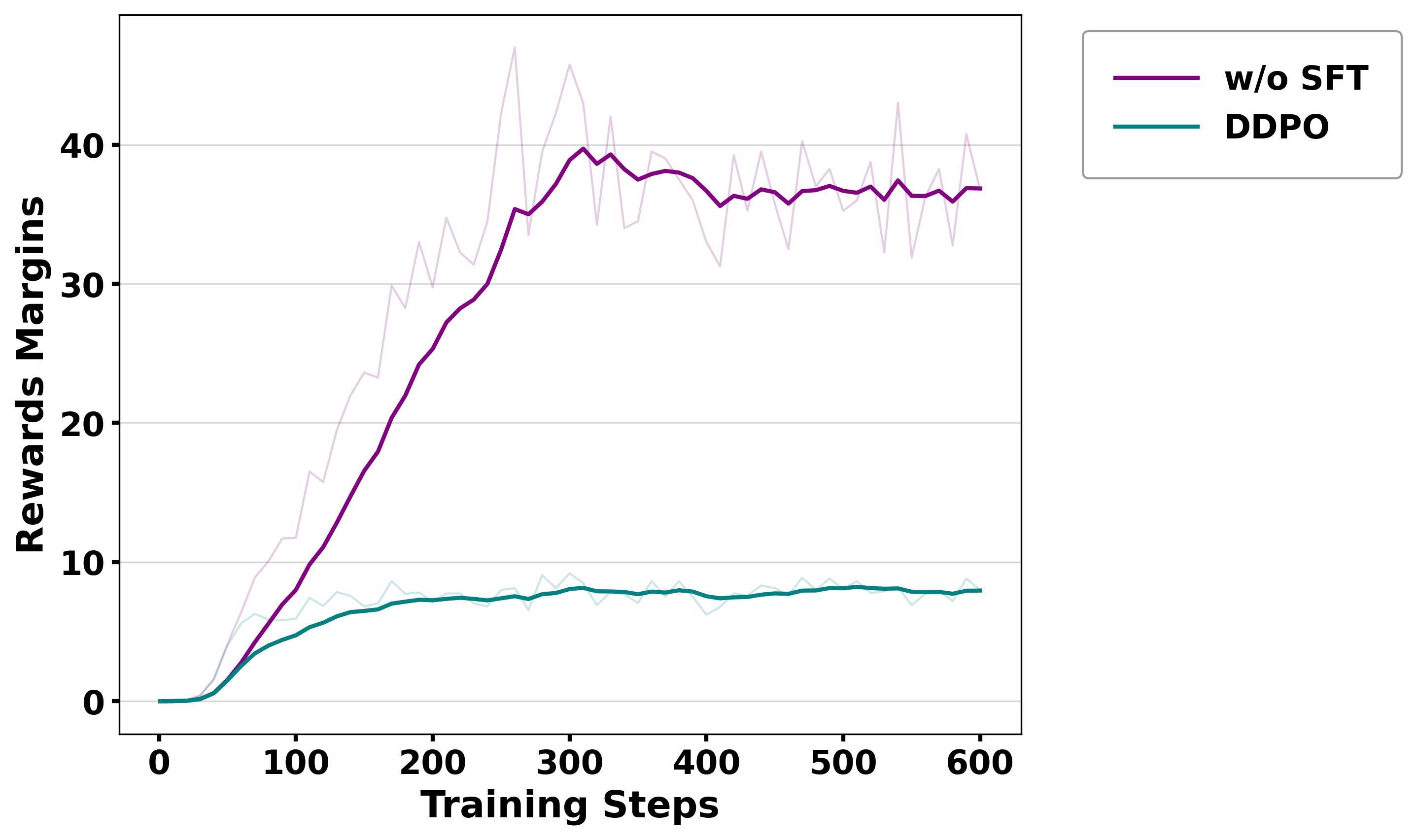}
        \caption{Reward/margin}
        \label{fig:reward-margins}
    \end{subfigure}%
    \begin{subfigure}[b]{0.33\textwidth}
        \centering
        \includegraphics[width=\textwidth]{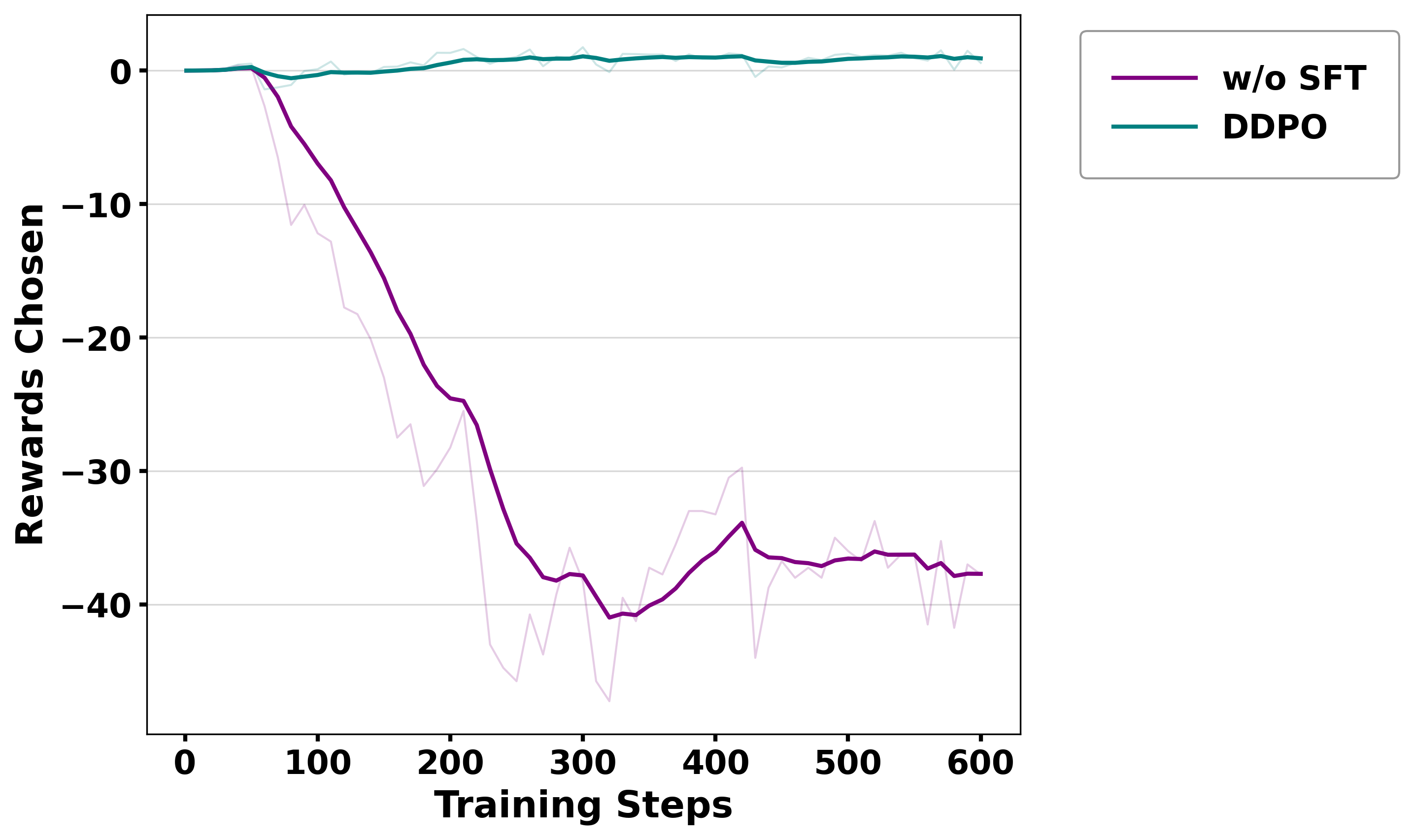}
        \caption{Reward/chosen}
        \label{fig:reward-chosen}
    \end{subfigure}
    \caption{Training curves with or without the SFT loss.}
    \label{fig:training_curves}
\end{figure*}

Table~\ref{tab:ablation-ddpo-appendix} extends the optimization ablation in Table~\ref{tab:ablation-ddpo} to the same two additional backbones. The results follow the same broad pattern as the results for the 8B model. DDPO generally improves over SFT, while removing token weighting or the SFT term weakens the overall profile. These trends suggest that the optimization design is not specific to the main backbone.

\subsection{Training Curves}\label{app:training_curves}

Figure~\ref{fig:training_curves} compares the full DDPO objective with the variant that removes the SFT term. Without SFT regularization, the DPO margin increases, but the implicit rewards for both chosen and rejected responses decrease. This indicates that the model can improve relative preference separation while drifting away from the refined scaffold distribution. With the SFT term, the margin grows more moderately and the reward for chosen responses remains better anchored. These trends support the main-text observation that SFT regularization helps stabilize preference optimization.

\subsection{Computational Efficiency}\label{app:computational_efficiency}

Data generation uses GPT-4o-mini for targeted refinement. Filtering and refining 2,401 valid training pairs from the original 19,938 entries costs about \$20 and takes 8--10 hours, depending on API latency. This is a one-time preprocessing cost and does not affect later model use.

We train parameter-efficient LoRA adapters with DeepSpeed-Zero3 on a single A800-40G GPU. The objective introduces no additional computational complexity over other DPO-based methods such as DDR. Training takes less than one hour for 3B-parameter models and less than four hours for 14B-parameter models. Scaffolds are used only for training supervision, so inference has the same latency and computational cost as Vanilla RAG.

\begin{table*}[]
\centering
\resizebox{\textwidth}{!}{
\begin{tabular}{lcccccccccccc}
\toprule
\multicolumn{1}{c}{\multirow{3}{*}{\textbf{Optimization Method}}} & \multicolumn{2}{c}{\textbf{In Domain}} & \multicolumn{10}{c}{\textbf{Out Of Domain}} \\ 
\cmidrule(lr){2-3} \cmidrule(lr){4-13}
\multicolumn{1}{c}{} & \multicolumn{2}{c}{\textbf{Musique}} & \multicolumn{2}{c}{\textbf{NQ}} & \multicolumn{2}{c}{\textbf{HotpotQA}} & \multicolumn{2}{c}{\textbf{PopQA}} & \multicolumn{2}{c}{\textbf{TruthfulQA}} & \multicolumn{2}{c}{\textbf{Zero-shot RE}} \\
\cmidrule(lr){2-3} \cmidrule(lr){4-5} \cmidrule(lr){6-7} \cmidrule(lr){8-9} \cmidrule(lr){10-11} \cmidrule(lr){12-13}
\multicolumn{1}{c}{} & \textbf{EM} & \textbf{F1} & \textbf{EM} & \textbf{F1} & \textbf{EM} & \textbf{F1} & \textbf{EM} & \textbf{F1} & \textbf{BLEU} & \textbf{Rouge-1} & \textbf{F1} & \textbf{Precision} \\ 
\midrule

% ======================= Llama-3.2-3B =======================
\multicolumn{13}{l}{\textit{Llama-3.2-3B-Instruct}} \\ 
\midrule
None (Vanilla RAG) & 4.47 & 9.89 & 33.20 & 45.59 & \textbf{27.89} & 37.75 & 34.32 & 41.60 & 6.36 & 13.78 & 48.70 & 47.40 \\
SFT & 2.73 & 7.29 & 20.06 & 30.46 & 21.73 & 31.35 & 29.52 & 35.80 & 5.96 & 14.02 & 25.32 & 22.83 \\
KARE (DDPO) & \textbf{5.59} & 13.28 & \textbf{33.21} & \textbf{46.56} & 27.62 & \textbf{39.59} & \textbf{38.29} & \textbf{45.52} & 7.58 & 17.75 & \textbf{51.94} & \textbf{49.03} \\
\quad - w/o SFT Loss & 4.10 & \textbf{13.99} & 22.32 & 38.37 & 23.13 & 37.50 & 31.91 & 41.57 & 7.96 & \textbf{20.64} & 49.36 & 44.64 \\
\quad - w/o Token Weight & 4.63 & 13.97 & 26.40 & 41.95 & 24.27 & 38.53 & 33.32 & 42.87 & \textbf{7.97} & 19.16 & 51.20 & 48.32 \\ 
\midrule

% ======================= Qwen2.5-14B =======================
\multicolumn{13}{l}{\textit{Qwen2.5-14B-Instruct}} \\ 
\midrule
None (Vanilla RAG) & 6.66 & 14.58 & 32.67 & 47.01 & 30.64 & 42.31 & 36.38 & 43.98 & 6.90 & 18.15 & 53.26 & 49.27 \\
SFT & 6.08 & 14.12 & 32.23 & 45.70 & 28.32 & 40.56 & 36.57 & 43.86 & 8.32 & 20.46 & 51.72 & 49.27 \\
KARE (DDPO) & \textbf{8.32} & \textbf{17.48} & \textbf{36.19} & \textbf{49.96} & \textbf{33.96} & \textbf{46.45} & \textbf{38.58} & \textbf{46.09} & 8.80 & 20.39 & \textbf{54.47} & \textbf{50.55} \\
\quad - w/o SFT Loss & 6.29 & 15.70 & 25.18 & 41.25 & 25.18 & 41.25 & 33.42 & 42.03 & 8.54 & \textbf{21.95} & 52.10 & 47.22 \\
\quad - w/o Token Weight & 7.57 & 16.58 & 34.99 & 49.00 & 32.84 & 45.47 & 38.12 & 45.65 & \textbf{9.00} & 20.17 & 54.07 & 50.13 \\ 
\bottomrule
\end{tabular}
}
\caption{Ablation analysis of optimization methods across varying model scales. We compare the SFT baseline, the DDPO method, and its ablation variants (w/o SFT Loss, w/o Token Weighting) on Llama-3.2-3B-Instruct and Qwen2.5-14B-Instruct.}
\label{tab:ablation-ddpo-appendix}
\end{table*}

\subsection{Scaffold Data Quality Analysis}\label{app:scaffold_data_quality}

Table~\ref{tab:data_funnel} traces the 19,938 Musique training examples through the construction pipeline. KARE retains cases in which the retrieved documents contain enough evidence but the base generator answers incorrectly. Gold answers are used only during offline construction to check evidence adequacy and refinement success, and are not available during inference. Most examples are removed because the retrieved documents do not contain enough evidence. The final set contains 2,401 preference pairs.

\begin{center}
\centering
\resizebox{\columnwidth}{!}{
\begin{tabular}{lrr}
\toprule
\textbf{Stage} & \textbf{Count} & \textbf{Ratio} \\
\midrule
Original examples & 19,938 & 100.0\% \\
Base generator correct & 3,735 & 18.7\% \\
Insufficient retrieved evidence & 13,216 & 66.3\% \\
Refinement/verification failure & 586 & 2.9\% \\
Final preference pairs & 2,401 & 12.0\% \\
\bottomrule
\end{tabular}
}
{\captionsetup{hypcap=false}
\captionof{table}{Filtering statistics for the default Musique and GPT-4o-mini data-construction pipeline. Ratios are relative to the original training set.}
\label{tab:data_funnel}}
\end{center}

Of the 586 refinement or verification failures, 573 result from truncation at the 256-token scaffold limit, 7 are formatting errors, 4 contain an insufficient evidence chain, and 2 still lead to an incorrect answer after refinement. We tested a 512-token limit, but it yielded few additional valid pairs while increasing data-construction and training costs. We therefore retain the 256-token limit. This breakdown is specific to the default Musique and GPT-4o-mini pipeline and may differ for other expert models or source datasets.

We manually inspect 100 sampled refined instances. A scaffold is valid if its corrected evidence structure is sufficient to derive the gold answer. Of the 100 samples, 97 are fully valid, 2 are partially valid, and 1 is invalid. The two partially valid samples capture the main entity or reasoning chain but lack the detail needed for the final answer. We also check whether the retrieved documents fully support each chosen scaffold and find two unsupported cases among the 97 initially valid samples. Under this stricter criterion, 95 of the 100 sampled instances remain fully valid.

We next evaluate whether data quality affects training. Starting from the refined training set, we randomly corrupt 15\% and 30\% of the scaffolds by deleting answer-supporting edges, replacing edges or relations, and adding distractor evidence.

Table~\ref{tab:scaffold_corruption} shows that the clean KARE setting achieves the strongest results. Performance decreases under 15\% corruption and drops further under 30\% corruption. Although the corrupted variants still remain above Vanilla RAG, the consistent degradation indicates that scaffold quality materially affects the training signal. These results support the need for a strict refinement and filtering pipeline when constructing KARE training data.

\begin{table*}[]
\centering
\resizebox{\textwidth}{!}{
\begin{tabular}{lcccccccccccc}
\toprule
\multicolumn{1}{c}{\multirow{3}{*}{\textbf{Training Setting}}} & \multicolumn{2}{c}{\textbf{In Domain}} & \multicolumn{10}{c}{\textbf{Out Of Domain}} \\ 
\cmidrule(lr){2-3} \cmidrule(lr){4-13}
\multicolumn{1}{c}{} & \multicolumn{2}{c}{\textbf{Musique}} & \multicolumn{2}{c}{\textbf{NQ}} & \multicolumn{2}{c}{\textbf{HotpotQA}} & \multicolumn{2}{c}{\textbf{PopQA}} & \multicolumn{2}{c}{\textbf{TruthfulQA}} & \multicolumn{2}{c}{\textbf{Zero-shot RE}} \\
\cmidrule(lr){2-3} \cmidrule(lr){4-5} \cmidrule(lr){6-7} \cmidrule(lr){8-9} \cmidrule(lr){10-11} \cmidrule(lr){12-13}
\multicolumn{1}{c}{} & \textbf{EM} & \textbf{F1} & \textbf{EM} & \textbf{F1} & \textbf{EM} & \textbf{F1} & \textbf{EM} & \textbf{F1} & \textbf{BLEU} & \textbf{Rouge-1} & \textbf{F1} & \textbf{Precision} \\ 
\midrule
None (Vanilla RAG) & 6.00 & 12.37 & 34.64 & 48.30 & 29.52 & 40.66 & 35.43 & 44.10 & 5.43 & 15.03 & 51.62 & 49.30 \\
KARE & \textbf{8.02} & \textbf{15.75} & \textbf{37.86} & \textbf{50.84} & \textbf{32.36} & \textbf{44.29} & \textbf{40.88} & \textbf{47.77} & \textbf{7.42} & \textbf{17.76} & \textbf{59.52} & \textbf{58.00} \\
\quad + 15\% corrupted scaffolds & 7.53 & 15.17 & 37.15 & 50.02 & 32.33 & 43.72 & 39.41 & 46.84 & 7.09 & 17.62 & 58.99 & 56.68 \\
\quad + 30\% corrupted scaffolds & 6.83 & 14.71 & 36.53 & 49.37 & 31.47 & 43.27 & 38.52 & 46.27 & 6.83 & 16.94 & 58.44 & 55.19 \\ 
\bottomrule
\end{tabular}
}
\caption{Effect of scaffold data corruption during training. Corruption is applied to a random subset of training scaffolds by deleting answer-supporting edges, replacing edges or relations, and adding distractor evidence.}
\label{tab:scaffold_corruption}
\end{table*}

\subsection{Transfer Across Retrieval Backends}\label{app:retriever_transfer}

KARE is trained using data constructed with bge-large-en-v1.5 retrieval. To examine whether the learned generator depends on this retrieval backend, we evaluate the same trained model with e5-base-v2\cite{wang2022text} and BM25 at inference time. No additional training is performed for these retrieval settings.

As shown in Table~\ref{tab:retriever_transfer}, KARE improves over the no-training baseline under all three retrieval backends. The gains persist when the inference retriever is changed to e5-base-v2 or BM25, although the absolute scores depend on retrieval quality. This result suggests that the training signal is not limited to the bge-large-en-v1.5 retrieval distribution. It can also improve evidence use under unseen retrieval backends.

\begin{table*}[]
\centering
\resizebox{\textwidth}{!}{
\begin{tabular}{lcccccccccccc}
\toprule
\multicolumn{1}{c}{\multirow{3}{*}{\textbf{Train Method}}} & \multicolumn{2}{c}{\textbf{In Domain}} & \multicolumn{10}{c}{\textbf{Out Of Domain}} \\ 
\cmidrule(lr){2-3} \cmidrule(lr){4-13}
\multicolumn{1}{c}{} & \multicolumn{2}{c}{\textbf{Musique}} & \multicolumn{2}{c}{\textbf{NQ}} & \multicolumn{2}{c}{\textbf{HotpotQA}} & \multicolumn{2}{c}{\textbf{PopQA}} & \multicolumn{2}{c}{\textbf{TruthfulQA}} & \multicolumn{2}{c}{\textbf{Zero-shot RE}} \\
\cmidrule(lr){2-3} \cmidrule(lr){4-5} \cmidrule(lr){6-7} \cmidrule(lr){8-9} \cmidrule(lr){10-11} \cmidrule(lr){12-13}
\multicolumn{1}{c}{} & \textbf{EM} & \textbf{F1} & \textbf{EM} & \textbf{F1} & \textbf{EM} & \textbf{F1} & \textbf{EM} & \textbf{F1} & \textbf{BLEU} & \textbf{Rouge-1} & \textbf{F1} & \textbf{Precision} \\ 
\midrule

\multicolumn{13}{l}{\textit{bge-large-en-v1.5}} \\
\midrule
None & 6.00 & 12.37 & 34.64 & 48.30 & 29.52 & 40.66 & 35.43 & 44.10 & 5.43 & 15.03 & 51.62 & 49.30 \\
KARE & \textbf{8.02} & \textbf{15.75} & \textbf{37.86} & \textbf{50.84} & \textbf{32.36} & \textbf{44.29} & \textbf{40.88} & \textbf{47.77} & \textbf{7.42} & \textbf{17.76} & \textbf{59.52} & \textbf{58.00} \\
\midrule

\multicolumn{13}{l}{\textit{e5-base-v2}} \\
\midrule
None & 4.85 & 11.40 & 32.88 & 45.90 & 27.52 & 39.36 & 37.31 & 47.29 & 4.63 & 17.20 & 56.78 & 53.50 \\
KARE & \textbf{6.73} & \textbf{15.15} & \textbf{34.02} & \textbf{46.75} & \textbf{29.86} & \textbf{43.49} & \textbf{42.88} & \textbf{50.83} & \textbf{6.58} & \textbf{21.20} & \textbf{61.33} & \textbf{60.97} \\
\midrule

\multicolumn{13}{l}{\textit{BM25}} \\
\midrule
None & 2.85 & 7.56 & 16.54 & 23.93 & 19.81 & 28.86 & 15.60 & 20.04 & 7.77 & 22.27 & 38.18 & 36.01 \\
KARE & \textbf{3.73} & \textbf{9.28} & \textbf{18.98} & \textbf{27.86} & \textbf{22.15} & \textbf{31.64} & \textbf{17.99} & \textbf{22.36} & \textbf{11.27} & \textbf{26.42} & \textbf{42.95} & \textbf{43.41} \\
\bottomrule
\end{tabular}
}
\caption{Retriever transfer results on Llama-3.1-8B-Instruct. KARE is trained with bge-large-en-v1.5 retrieval and evaluated with bge-large-en-v1.5, e5-base-v2, and BM25 retrieval at inference time.}
\label{tab:retriever_transfer}
\end{table*}

\subsection{Domain Shift on RAGBench}\label{app:domain_shift}

We further evaluate domain shift using three RAGBench domains\cite{friel2024ragbench}. CUAD covers legal QA, FinQA covers financial QA, and PubMedQA covers biomedical QA. We use the same KARE-trained Llama-3.1-8B-Instruct model from the main experiments, without domain-specific fine-tuning. We report F1 together with adherence and completeness, where the latter two metrics are judged by GPT-4.1-mini.

Table~\ref{tab:ragbench_domain_shift} in the main paper compares KARE with Vanilla RAG, DDR, and RAFT under this protocol. KARE obtains the best result on all six adherence and completeness metrics and on eight of nine metrics overall; RAFT obtains the best CUAD F1. These results are limited to the three evaluated domains and this fixed-retriever, generator-only setup.

\subsection{Generalization Across Different Generation Pipelines}\label{app:pipeline_generalization}

We evaluate KARE with five generation pipelines: Vanilla RAG, Chain-of-Thought\cite{wei2022chain}, Chain-of-Note\cite{yu-etal-2024-chain}, StructRAG\cite{li2025structrag}, and IRCoT\cite{trivedi2022interleaving}. Each pipeline is compared with its no-training baseline. IRCoT uses the same task metrics as the other pipelines rather than a separate accuracy-only report.

\begin{table*}[]
\centering
\resizebox{\textwidth}{!}{
\begin{tabular}{llcccccccccccc}
\toprule
\multicolumn{1}{c}{\multirow{3}{*}{\textbf{\begin{tabular}[c]{@{}c@{}}Inference\\ Pipeline\end{tabular}}}} & \multicolumn{1}{c}{\multirow{3}{*}{\textbf{\begin{tabular}[c]{@{}c@{}}Train\\ Method\end{tabular}}}} & \multicolumn{2}{c}{\textbf{In Domain}} & \multicolumn{10}{c}{\textbf{Out Of Domain}} \\ 
\cmidrule(lr){3-4} \cmidrule(lr){5-14}
\multicolumn{1}{c}{} & \multicolumn{1}{c}{} & \multicolumn{2}{c}{\textbf{Musique}} & \multicolumn{2}{c}{\textbf{NQ}} & \multicolumn{2}{c}{\textbf{HotpotQA}} & \multicolumn{2}{c}{\textbf{PopQA}} & \multicolumn{2}{c}{\textbf{TruthfulQA}} & \multicolumn{2}{c}{\textbf{Zero-shot RE}} \\
\cmidrule(lr){3-4} \cmidrule(lr){5-6} \cmidrule(lr){7-8} \cmidrule(lr){9-10} \cmidrule(lr){11-12} \cmidrule(lr){13-14}
\multicolumn{1}{c}{} & \multicolumn{1}{c}{} & \textbf{EM} & \textbf{F1} & \textbf{EM} & \textbf{F1} & \textbf{EM} & \textbf{F1} & \textbf{EM} & \textbf{F1} & \textbf{BLEU} & \textbf{Rouge-1} & \textbf{F1} & \textbf{Precision} \\ 
\midrule
\multirow{2}{*}{Vanilla} & None & 6.00 & 12.37 & 34.64 & 48.30 & 29.52 & 40.66 & 35.43 & 44.10 & 5.43 & 15.03 & 51.62 & 49.30 \\
 & KARE & \textbf{8.02} & \textbf{15.75} & \textbf{37.86} & \textbf{50.84} & \textbf{32.36} & \textbf{44.29} & \textbf{40.88} & \textbf{47.77} & \textbf{7.42} & \textbf{17.76} & \textbf{59.52} & \textbf{58.00} \\ 
\hdashline
\multirow{2}{*}{CoT} & None & 8.52 & 16.40 & 35.63 & 48.48 & 31.11 & 42.45 & 37.71 & 44.35 & 3.41 & 10.27 & 51.26 & 48.61 \\
 & KARE & \textbf{10.34} & \textbf{19.28} & \textbf{37.22} & \textbf{50.28} & \textbf{34.40} & \textbf{46.82} & \textbf{40.38} & \textbf{46.96} & \textbf{4.51} & \textbf{11.35} & \textbf{54.44} & \textbf{52.41} \\ 
\hdashline
\multirow{2}{*}{CoN} & None & 6.83 & 14.47 & 35.06 & 48.61 & 29.05 & 40.70 & 36.06 & 43.32 & 5.81 & 11.96 & 45.60 & 42.80 \\
 & KARE & \textbf{9.18} & \textbf{18.44} & \textbf{37.55} & \textbf{50.73} & \textbf{33.11} & \textbf{45.34} & \textbf{39.36} & \textbf{46.30} & \textbf{7.14} & \textbf{12.83} & \textbf{49.41} & \textbf{48.04} \\ 
\hdashline
\multirow{2}{*}{StructRAG} & None & 3.97 & 10.30 & 26.04 & 37.94 & 18.27 & 29.96 & 30.83 & 30.59 & 9.75 & 25.44 & 42.43 & 38.93 \\
 & KARE & \textbf{8.07} & \textbf{16.69} & \textbf{29.14} & \textbf{40.86} & \textbf{21.27} & \textbf{34.13} & \textbf{33.50} & \textbf{39.00} & \textbf{11.32} & \textbf{27.91} & \textbf{45.56} & \textbf{42.44} \\ 
\hdashline
\multirow{2}{*}{IRCoT} & None & 9.97 & 18.58 & 34.13 & 46.63 & 36.10 & 49.36 & 38.42 & 46.97 & 4.80 & 15.81 & 54.95 & 51.71 \\
 & KARE & \textbf{13.90} & \textbf{23.14} & \textbf{35.18} & \textbf{48.89} & \textbf{38.24} & \textbf{51.86} & \textbf{41.26} & \textbf{48.85} & \textbf{6.61} & \textbf{20.47} & \textbf{58.66} & \textbf{55.56} \\ 
\bottomrule
\end{tabular}
}
\caption{Generalization across inference pipelines on Llama-3.1-8B-Instruct. We compare no additional training with KARE under Vanilla RAG, Chain-of-Thought (CoT), Chain-of-Note (CoN), StructRAG, and IRCoT.}
\label{tab:pipeline-all}
\end{table*}

Table~\ref{tab:pipeline-all} shows that KARE improves every reported metric under each of the five pipelines. The gains therefore do not depend on the Vanilla RAG prompt used in the main comparison.

\subsection{Additional Inference-Time Comparisons}\label{app:inference_comparisons}

The main experiments use one-pass direct-answer inference so that KARE retains the same interface as Vanilla RAG. This section examines three alternatives that change the inference procedure: explicit scaffold generation, context compression, and Self-RAG-style reflection. We report them separately because they do not share the main comparison's inference protocol.

\paragraph{Explicit scaffold inference.}
Although KARE is trained on scaffold pairs, it does not generate a scaffold at test time by default. We also evaluate a two-step setting in which the trained model first constructs a scaffold and then answers from it. Table~\ref{tab:explicit_scaffold_inference} shows that explicit scaffold generation improves MuSiQue and HotpotQA F1 but lowers the other four reported metrics. It also requires separate scaffold and answer generation steps. We therefore retain direct answering as the default and treat explicit scaffold inference as a task-dependent option for compositional QA.

\begin{table*}[t]
\centering
\resizebox{0.9\textwidth}{!}{
\begin{tabular}{lcccccc}
\toprule
\textbf{Inference Setting} & \textbf{MuSiQue F1} & \textbf{NQ F1} & \textbf{HotpotQA F1} & \textbf{PopQA F1} & \textbf{TQA R-1} & \textbf{ZSRE F1} \\
\midrule
Direct KARE & 15.75 & \textbf{50.84} & 44.29 & \textbf{47.77} & \textbf{17.76} & \textbf{59.52} \\
KARE, Scaffold-to-Answer & \textbf{20.53} & 49.36 & \textbf{46.01} & 47.55 & 12.92 & 37.12 \\
\bottomrule
\end{tabular}
}
\caption{Direct-answer and explicit Scaffold-to-Answer inference with the same KARE-trained Llama-3.1-8B-Instruct generator. TQA R-1 denotes TruthfulQA ROUGE-1, and ZSRE denotes zero-shot relation extraction.}
\label{tab:explicit_scaffold_inference}
\end{table*}

\paragraph{Context compression.}
We use LLMLingua-2~\citep{pan-etal-2024-llmlingua} to compress the retrieved contexts before generation, with target retained-context ratios of 50\% and 30\%. Table~\ref{tab:llmlingua_compression} compares compression with both Vanilla RAG and KARE. Relative to uncompressed Vanilla RAG, LLMLingua-2 lowers seven of the twelve metrics at 50\% and eleven at 30\%. Under the same compressed inputs, KARE exceeds the corresponding Vanilla setting on all twelve metrics at 50\% and eleven at 30\%. These results do not support context compression as a substitute for generator training in this setting, but show that KARE can be used with compressed contexts.

\begin{table*}[t]
\centering
\resizebox{\textwidth}{!}{
\begin{tabular}{lcccccccccccc}
\toprule
\multicolumn{1}{c}{\multirow{3}{*}{\textbf{Method}}} & \multicolumn{2}{c}{\textbf{In Domain}} & \multicolumn{10}{c}{\textbf{Out Of Domain}} \\
\cmidrule(lr){2-3} \cmidrule(lr){4-13}
& \multicolumn{2}{c}{\textbf{MuSiQue}} & \multicolumn{2}{c}{\textbf{NQ}} & \multicolumn{2}{c}{\textbf{HotpotQA}} & \multicolumn{2}{c}{\textbf{PopQA}} & \multicolumn{2}{c}{\textbf{TruthfulQA}} & \multicolumn{2}{c}{\textbf{Zero-shot RE}} \\
\cmidrule(lr){2-3} \cmidrule(lr){4-5} \cmidrule(lr){6-7} \cmidrule(lr){8-9} \cmidrule(lr){10-11} \cmidrule(lr){12-13}
& \textbf{EM} & \textbf{F1} & \textbf{EM} & \textbf{F1} & \textbf{EM} & \textbf{F1} & \textbf{EM} & \textbf{F1} & \textbf{BLEU} & \textbf{R-1} & \textbf{F1} & \textbf{Prec.} \\
\midrule
Vanilla RAG & 6.00 & 12.37 & 34.64 & 48.30 & 29.52 & 40.66 & 35.43 & 44.10 & 5.43 & 15.03 & 51.62 & 49.30 \\
\quad + LLMLingua-2 (50\%) & 5.71 & 12.63 & 31.50 & 44.26 & 28.12 & 40.11 & 34.84 & 42.20 & 6.17 & 16.24 & 55.14 & 53.24 \\
\quad + LLMLingua-2 (30\%) & 4.51 & 11.16 & 30.08 & 41.88 & 24.17 & 34.90 & 31.31 & 37.09 & 6.93 & 14.90 & 44.16 & 43.76 \\
KARE & \textbf{8.02} & \textbf{15.75} & \textbf{37.86} & \textbf{50.84} & \textbf{32.36} & \textbf{44.29} & \textbf{40.88} & \textbf{47.77} & 7.42 & \textbf{17.76} & 59.52 & \textbf{58.00} \\
\quad + LLMLingua-2 (50\%) & 6.29 & 14.65 & 33.35 & 45.39 & 29.63 & 41.96 & 38.05 & 44.71 & 7.65 & 16.59 & \textbf{59.56} & 57.63 \\
\quad + LLMLingua-2 (30\%) & 5.54 & 13.04 & 32.69 & 44.03 & 26.10 & 37.67 & 34.91 & 40.47 & \textbf{7.70} & 14.41 & 49.87 & 49.13 \\
\bottomrule
\end{tabular}
}
\caption{Context-compression results on Llama-3.1-8B-Instruct. Percentages denote the target fraction of the retrieved context retained by LLMLingua-2. KARE uses its default direct-answer inference in all rows.}
\label{tab:llmlingua_compression}
\end{table*}

\paragraph{Self-RAG adaptations.}
Native Self-RAG~\citep{asai2023self} uses adaptive retrieval and reflection-token decoding, which cannot be fully preserved in our fixed-retriever protocol. We therefore report two controlled adaptations rather than a reproduction of the complete system. Zero-shot Self-RAG scores candidates using prompted relevance, support, and utility labels without training reflection tokens. Self-RAG-LoRA trains the Llama-3.1-8B-Instruct backbone on all 145,533 decontaminated public examples and uses learned reflection-token probabilities for candidate scoring. Table~\ref{tab:selfrag_adaptations} shows that Self-RAG-LoRA obtains the best NQ EM and TruthfulQA ROUGE-1, while KARE obtains the best result on the other ten metrics. These results describe the controlled fixed-retriever setting and should not be read as a comparison with native Self-RAG under its original inference pipeline.

\begin{table*}[t]
\centering
\resizebox{\textwidth}{!}{
\begin{tabular}{lcccccccccccc}
\toprule
\multicolumn{1}{c}{\multirow{3}{*}{\textbf{Method}}} & \multicolumn{2}{c}{\textbf{In Domain}} & \multicolumn{10}{c}{\textbf{Out Of Domain}} \\
\cmidrule(lr){2-3} \cmidrule(lr){4-13}
& \multicolumn{2}{c}{\textbf{MuSiQue}} & \multicolumn{2}{c}{\textbf{NQ}} & \multicolumn{2}{c}{\textbf{HotpotQA}} & \multicolumn{2}{c}{\textbf{PopQA}} & \multicolumn{2}{c}{\textbf{TruthfulQA}} & \multicolumn{2}{c}{\textbf{Zero-shot RE}} \\
\cmidrule(lr){2-3} \cmidrule(lr){4-5} \cmidrule(lr){6-7} \cmidrule(lr){8-9} \cmidrule(lr){10-11} \cmidrule(lr){12-13}
& \textbf{EM} & \textbf{F1} & \textbf{EM} & \textbf{F1} & \textbf{EM} & \textbf{F1} & \textbf{EM} & \textbf{F1} & \textbf{BLEU} & \textbf{R-1} & \textbf{F1} & \textbf{Prec.} \\
\midrule
Vanilla RAG & 6.00 & 12.37 & 34.64 & 48.30 & 29.52 & 40.66 & 35.43 & 44.10 & 5.43 & 15.03 & 51.62 & 49.30 \\
Zero-shot Self-RAG & 5.83 & 11.61 & 33.13 & 42.79 & 28.83 & 38.02 & 33.12 & 38.10 & 5.83 & 11.46 & 42.57 & 42.22 \\
Self-RAG-LoRA (145,533) & 1.12 & 6.13 & \textbf{40.89} & 48.58 & 8.85 & 19.62 & 2.50 & 17.42 & 5.81 & \textbf{25.86} & 46.25 & 44.90 \\
KARE & \textbf{8.02} & \textbf{15.75} & 37.86 & \textbf{50.84} & \textbf{32.36} & \textbf{44.29} & \textbf{40.88} & \textbf{47.77} & \textbf{7.42} & 17.76 & \textbf{59.52} & \textbf{58.00} \\
\bottomrule
\end{tabular}
}
\caption{Controlled Self-RAG adaptations on Llama-3.1-8B-Instruct with the same fixed retriever used in the KARE experiments. The Self-RAG-LoRA row uses the full decontaminated public training set.}
\label{tab:selfrag_adaptations}
\end{table*}

\subsection{Impact of Training Dataset Selection}\label{app:dataset_selection}

We selected Musique as the primary training dataset because it emphasizes compositional multi-hop reasoning and fragmented evidence. These properties make it well suited for constructing scaffolds that test evidence organization under retrieval noise.

To verify that KARE is not dependent on the specific distribution of the Musique dataset, we conduct additional experiments using HotpotQA and Natural Questions (NQ) as training sources. This appendix uses Qwen2.5-14B-Instruct as the refinement expert for all source datasets, including the Musique row in Table~\ref{tab:cross_dataset_generalization}. Under this common-expert setting, we generate 8,500 training pairs from HotpotQA and 9,008 pairs from NQ.

Since the main experiments use GPT-4o-mini as the default expert, the absolute scores in this appendix are not directly comparable to the main GPT-4o-mini setting. The comparison examines whether KARE remains effective with different source datasets under a shared refinement expert. Because the number of accepted pairs differs across datasets, it is not intended to isolate source-dataset effects under a matched data budget. As shown in Table~\ref{tab:cross_dataset_generalization}, models trained on different source datasets improve over Vanilla RAG across multiple benchmarks, suggesting that KARE is not tied to a single training distribution.

\begin{table*}[]
\centering
\resizebox{\textwidth}{!}{
\begin{tabular}{lcccccccccccc}
\toprule
\multicolumn{1}{c}{\multirow{2}{*}{\textbf{Training Dataset}}} & \multicolumn{2}{c}{\textbf{Musique}} & \multicolumn{2}{c}{\textbf{NQ}} & \multicolumn{2}{c}{\textbf{HotpotQA}} & \multicolumn{2}{c}{\textbf{PopQA}} & \multicolumn{2}{c}{\textbf{TruthfulQA}} & \multicolumn{2}{c}{\textbf{Zero-shot RE}} \\ 
\cmidrule(lr){2-3} \cmidrule(lr){4-5} \cmidrule(lr){6-7} \cmidrule(lr){8-9} \cmidrule(lr){10-11} \cmidrule(lr){12-13}
\multicolumn{1}{c}{} & \textbf{EM} & \textbf{F1} & \textbf{EM} & \textbf{F1} & \textbf{EM} & \textbf{F1} & \textbf{EM} & \textbf{F1} & \textbf{BLEU} & \textbf{Rouge-1} & \textbf{F1} & \textbf{Precision} \\ 
\midrule
None (Vanilla RAG) & 6.00 & 12.37 & 34.64 & 48.30 & 29.52 & 40.66 & 35.43 & 44.10 & 5.43 & 15.03 & 51.62 & 49.30 \\
Natural Questions & 5.88 & 12.73 & 37.36 & 49.97 & 30.44 & 41.20 & 38.48 & 45.71 & 7.42 & 17.80 & \textbf{59.57} & \textbf{57.79} \\
HotpotQA & 6.37 & 13.80 & \textbf{37.95} & \textbf{50.72} & \textbf{32.13} & 42.75 & \textbf{39.59} & \textbf{46.36} & \textbf{7.68} & 15.85 & 59.27 & 57.60 \\
Musique & \textbf{6.95} & \textbf{15.16} & 36.35 & 49.17 & 31.48 & \textbf{43.41} & 38.61 & 45.92 & 7.35 & \textbf{18.05} & 55.16 & 53.07 \\ 
\bottomrule
\end{tabular}
}
\caption{Performance comparison of KARE when trained on different datasets (Natural Questions, HotpotQA, and Musique). Llama-3.1-8B-Instruct serves as the generator backbone, while Qwen2.5-14B-Instruct is employed as the expert model for training data refinement.}
\label{tab:cross_dataset_generalization}
\end{table*}

\balance
\subsection{Case Study of Knowledge Refinement}\label{app:case_study}

To provide a qualitative analysis of the refinement mechanism, we present a detailed case study from the Musique dataset in Table~\ref{tab:case_study_box}. In this multi-hop reasoning scenario, the query asks for the creation year of the region containing Hertfordshire, which requires the model to filter out high-interference noise from retrieved documents. As observed in the negative sample ($s^{-}$), the base model successfully identifies relevant entities but fails to extract the specific temporal attribute associated with the ``East of England'' region, instead hallucinating an incorrect answer (``1963'') derived from a distractor document about ``Greater London.'' The expert refinement pipeline detects this logical gap and explicitly injects the missing evidence (\texttt{created\_in: 1994}) into the scaffold. This correction establishes a complete reasoning chain from the entity to the ground truth, demonstrating KARE's capability to mitigate retrieval noise and enhance factual fidelity through scaffolded refinement.

\begin{table*}[h]
\centering
\begin{tcolorbox}[colback=white, colframe=black, title=\textbf{Case Study: Refining Knowledge Structure for Multi-hop Reasoning}, boxrule=1pt, arc=2pt]

\textbf{Query:} \textit{What year saw the creation of the region where the county of Hertfordshire is located?}

\vspace{0.2cm}
\hrule
\vspace{0.2cm}

\textbf{Step 1: Retrieval Context (Selected Snippets)} \\
\small
\textbf{Doc 1 (Gold):} \textit{"East of England... It was \textcolor{mygreen}{\textbf{created in 1994}}... It includes the ceremonial counties of ... \textcolor{mygreen}{\textbf{Hertfordshire}}..."} \\
\textbf{Doc 5 (Distractor):} \textit{"Greater London... was first established... \textcolor{myred}{between 1963 and 1986}..."}

\vspace{0.2cm}
\hrule
\vspace{0.2cm}

\textbf{Step 2: Negative Sample ($s^{-}$) -- Generated by Base Model} \\
\small
The model successfully identifies entities but \textbf{fails to extract the creation year}, leading to a reasoning dead-end.

\begin{itemize}
    \setlength\itemsep{0em}
    \item \textbf{Entity: Hertfordshire} (Attributes: county, location: East of England)
    \item \textbf{Entity: East of England} (Attributes: region, population, NUTS classification) \textcolor{myred}{\textit{[Missing Attribute: Creation Date]}}
    \item \textbf{Entity: Greater London} (Attributes: established: 1963) $\leftarrow$ \textit{Distractor Information}
\end{itemize}
\textbf{Prediction from $s^{-}$:} \textcolor{myred}{1963} (Incorrect)

\vspace{0.2cm}
\hrule
\vspace{0.2cm}

\textbf{Step 3: Refined Positive Sample ($s^{+}$) -- After Expert Refinement} \\
\small
The refinement process locates the missing bridge in Doc 1 and structurally inserts the key temporal attribute.

\begin{itemize}
    \setlength\itemsep{0em}
    \item \textbf{Entity: Hertfordshire} (Attributes: county, location: East of England)
    \item \textbf{Entity: East of England} 
    \begin{itemize}
         \item Attributes: region, population: 5,847,000, \textcolor{mygreen}{\textbf{created\_in: 1994}}, includes: Hertfordshire
    \end{itemize}
\end{itemize}
\textbf{Prediction from $s^{+}$:} \textcolor{mygreen}{1994} (Correct)

\end{tcolorbox}
\caption{A comparison between the initial negative knowledge organization and the refined structure. The refinement explicitly captures the "created\_in" attribute that was previously missed, enabling the model to ignore the distractor date (1963) and answer correctly.}
\label{tab:case_study_box}
\end{table*}

\subsection{Experiment Prompts}\label{app:prompts}

This section presents all prompts used in our experiments. For Vanilla RAG, we use the standard prompt template provided by FlashRAG\cite{FlashRAG} without modification. Table~\ref{tab:prompt-graph} contains the prompts for Knowledge-Aware generation with a Knowledge Graph as the intermediate representation. Tables~\ref{tab:prompt-keypoints} and \ref{tab:prompt-summary} contain the prompts for Keypoints and Summary, respectively. Table~\ref{tab:prompt-data-construct} presents the prompts used during Knowledge Graph refinement.

\begin{table*}[h]
\centering
\small
\renewcommand{\arraystretch}{1.3}
\begin{tabularx}{\textwidth}{@{}p{2.5cm}X@{}}
\toprule
\textbf{Role} & \textbf{Content} \\ 

% ================= Stage 1 =================
\midrule
\multicolumn{2}{c}{\cellcolor{graybg}\textbf{Stage 1: Knowledge-Aware Extraction}} \\ 
\midrule

\textbf{System Prompt} & 
You are a helpful AI assistant distinctively skilled at extracting crucial information from documents helpful for answering a given question. For a given question, focus on identifying the key entities, their attributes, and their relationships that are directly relevant to generating an accurate answer. Only include entities and attributes that are crucial for understanding and forming the answer, and avoid unnecessary details.

\vspace{0.5em}
Your response \textbf{must include the following keys} and strictly \textbf{adhere to the exact structure} without any additional text before or after the keys:

\vspace{0.5em}
\texttt{Entities:}\newline
\texttt{- [Entity 1] (Attributes: [Attribute 1, Attribute 2, ...])}\newline
\texttt{- [Entity 2] (Attributes: [Attribute 1, Attribute 2, ...])}\newline
\texttt{...}

\vspace{0.5em}
\texttt{Relationships:}\newline
\texttt{1. [Entity 1] -> [Relationship] -> [Entity 2]}\newline
\texttt{2. ...}

\vspace{0.5em}
\textbf{Important Note:}\newline
1. \textbf{Do not include any extra commentary} or unnecessary details in the response.\newline
2. Closely follow the structure provided above and ensure that the response is concise and directly addresses the question.\newline
3. \textbf{Don't provide the answer to the question.} Instead, focus on extracting the key entities, attributes, and relationships that are essential for answering the question accurately. \\ 
\cmidrule(l){2-2}

\textbf{User Prompt} & 
Question: \texttt{\{question\}} \newline
Documents: \texttt{\{reference\}} \newline
Knowledge Graph: \\

% ================= Stage 2 =================
\midrule
\multicolumn{2}{c}{\cellcolor{graybg}\textbf{Stage 2: Chain-of-Thought Reasoning}} \\ 
\midrule

\textbf{System Prompt} & 
You are a helpful AI assistant good at reasoning on the knowledge graph to answer a given question. For a given question and relevant knowledge graph, provide the step-by-step reasoning process to derive the answer. Make sure the reasoning steps are logical, coherent, and directly related to the question and the information in the knowledge graph. \\ 
\cmidrule(l){2-2}
\textbf{User Prompt} & 
Question: \texttt{\{question\}} \newline
Knowledge Graph: \texttt{\{$s$\}} \newline
Reasoning Steps: \\

% ================= Stage 3 =================
\midrule
\multicolumn{2}{c}{\cellcolor{graybg}\textbf{Stage 3: Final Generation}} \\ 
\midrule

\textbf{System Prompt} & 
You are a helpful AI assistant good at generating final answers based on the reasoning steps provided for a given question.

\vspace{0.5em}
\textbf{Important Notes:}\newline
1. Make sure the final answer is accurate, concise, and directly addresses the question.\newline
2. Only give me the answer and \textbf{do not output any other words.} \\ 
\cmidrule(l){2-2}
\textbf{User Prompt} & 
Question: \texttt{\{question\}} \newline
Reasoning Steps: \texttt{\{$y_{\mathrm{CoT}}$\}} \newline
Answer: \\ 
\bottomrule

\end{tabularx}
\caption{Detailed Prompts for Knowledge-Aware Generation (Graph Format).}
\label{tab:prompt-graph}
\end{table*}

\begin{table*}[h]
\centering
\small
\renewcommand{\arraystretch}{1.3}
\begin{tabularx}{\textwidth}{@{}p{2.5cm}X@{}}
\toprule
\textbf{Role} & \textbf{Content} \\ 

% ================= Stage 1: Keypoints =================
\midrule
\multicolumn{2}{c}{\cellcolor{graybg}\textbf{Stage 1: Keypoints Extraction}} \\ 
\midrule

\textbf{System Prompt} & 
You are a helpful AI assistant good at extracting crucial information from documents which are helpful for answering a given question. For a given question, please thoroughly analyze the list of documents (given as strings) and extract the most relevant key points that directly answer the question. Ensure that each key point is directly supported by evidence found within the documents, and avoid unnecessary details.

\vspace{0.5em}
Your response \textbf{must include the following keys} and strictly \textbf{adhere to the exact structure} without any additional text before or after the keys:

\vspace{0.5em}
\texttt{Key Points:}\newline
\texttt{1. [Key Point 1]}\newline
\texttt{2. [Key Point 2]}\newline
\texttt{...}

\vspace{0.5em}
\textbf{Important Note:}\newline
1. \textbf{Do not include any extra commentary} or unnecessary details in the response.\newline
2. Closely follow the structure provided above and ensure that the response is concise and directly addresses the question.\newline
3. \textbf{Don't provide the answer to the question.} Instead, focus on extracting the key points that are essential for answering the question accurately. \\ 
\cmidrule(l){2-2}
\textbf{User Prompt} & 
Question: \texttt{\{question\}} \newline
Documents: \texttt{\{reference\}} \newline
KeyPoints: \\

% ================= Stage 2: CoT =================
\midrule
\multicolumn{2}{c}{\cellcolor{graybg}\textbf{Stage 2: Chain-of-Thought Reasoning}} \\ 
\midrule

\textbf{System Prompt} & 
You are a helpful AI assistant that is good at doing reasoning on the keypoints to answer a given question. For a given question and relevant keypoints, provide the step-by-step reasoning process to derive the answer. Make sure the reasoning steps are logical, coherent, and directly related to the question and the information in the keypoints. \\ 
\cmidrule(l){2-2}
\textbf{User Prompt} & 
Question: \texttt{\{question\}} \newline
Keypoints: \texttt{\{$y_{\mathrm{Key}}$\}} \newline
Reasoning Steps: \\

% ================= Stage 3: Generation =================
\midrule
\multicolumn{2}{c}{\cellcolor{graybg}\textbf{Stage 3: Final Generation}} \\ 
\midrule

\textbf{System Prompt} & 
You are a helpful AI assistant that is good at generating final answers based on the reasoning steps provided for a given question.

\vspace{0.5em}
\textbf{Important Notes:}\newline
1. Make sure the final answer is accurate, concise, and directly addresses the question.\newline
2. Only give me the answer and \textbf{do not output any other words.} \\ 
\cmidrule(l){2-2}
\textbf{User Prompt} & 
Question: \texttt{\{question\}} \newline
Reasoning Steps: \texttt{\{$y_{\mathrm{CoT}}$\}} \newline
Answer: \\ 
\bottomrule

\end{tabularx}
\caption{Detailed Prompts for Knowledge-Aware Generation (Keypoints Format).}
\label{tab:prompt-keypoints}
\end{table*}

\begin{table*}[h]
\centering
\small
\renewcommand{\arraystretch}{1.3}
\begin{tabularx}{\textwidth}{@{}p{2.5cm}X@{}}
\toprule
\textbf{Role} & \textbf{Content} \\ 

% ================= Stage 1: Summary Note =================
\midrule
\multicolumn{2}{c}{\cellcolor{graybg}\textbf{Stage 1: Summary Note Extraction}} \\ 
\midrule

\textbf{System Prompt} & 
You are a helpful AI assistant that is good at extracting crucial information from documents which are helpful for answering a given question. For a given question, you need to extract a note that is directly relevant to generating an accurate answer.

\vspace{0.5em}
\textbf{Important Note:}\newline
1. \textbf{Don't directly provide the answer to the question.} Instead, focus on extracting the relevant information that is essential for answering the question accurately and formulating a note.\newline
2. The length of the response is limited, so make sure to include only the most relevant information. \\ 
\cmidrule(l){2-2}
\textbf{User Prompt} & 
Question: \texttt{\{question\}} \newline
Documents: \texttt{\{reference\}} \newline
Note: \\

% ================= Stage 2: CoT =================
\midrule
\multicolumn{2}{c}{\cellcolor{graybg}\textbf{Stage 2: Chain-of-Thought Reasoning}} \\ 
\midrule

\textbf{System Prompt} & 
You are a helpful AI assistant that is good at doing reasoning on the note to answer a given question. For a given question with relevant note, provide the step-by-step reasoning process to derive the answer. Make sure the reasoning steps are logical, coherent, and directly related to the question and the information in the note. \\ 
\cmidrule(l){2-2}
\textbf{User Prompt} & 
Question: \texttt{\{question\}} \newline
Note: \texttt{\{$y_{\mathrm{Note}}$\}} \newline
Reasoning Steps: \\

% ================= Stage 3: Generation =================
\midrule
\multicolumn{2}{c}{\cellcolor{graybg}\textbf{Stage 3: Final Generation}} \\ 
\midrule

\textbf{System Prompt} & 
You are a helpful AI assistant that is good at generating final answers based on the reasoning steps provided for a given question.

\vspace{0.5em}
\textbf{Important Notes:}\newline
1. Make sure the final answer is accurate, concise, and directly addresses the question.\newline
2. Only give me the answer and \textbf{do not output any other words.} \\ 
\cmidrule(l){2-2}
\textbf{User Prompt} & 
Question: \texttt{\{question\}} \newline
Reasoning Steps: \texttt{\{$y_{\mathrm{CoT}}$\}} \newline
Answer: \\ 
\bottomrule

\end{tabularx}
\caption{Detailed Prompts for Knowledge-Aware Generation (Summary Format).}
\label{tab:prompt-summary}
\end{table*}

\begin{table*}[h]
\centering
\small
\renewcommand{\arraystretch}{1.3}
\begin{tabularx}{\textwidth}{@{}p{2.5cm}X@{}}
\toprule
\textbf{Role} & \textbf{Content} \\ 

% ================= Module 1 =================
\midrule
\multicolumn{2}{c}{\cellcolor{graybg}\textbf{Module: Check Answerability}} \\ 
\midrule

\textbf{System Prompt} & 
You are a helpful AI assistant very good at judging whether the answer can be derived from the documents for a given question. You will be given a question, a set of documents, and golden answers. Please determine whether any of the golden answers can be derived from the documents for the given question.

\vspace{0.5em}
If any of the golden answers can be derived from the documents, the judgement should be "True". If none of the golden answers can be derived from the documents, the judgement should be "False". Do not output any explanation, only output the judgement. \\ 
\cmidrule(l){2-2}
\textbf{User Prompt} & 
Question: \texttt{\{question\}} \newline
Documents: \texttt{\{reference\}} \newline
Golden Answers: \texttt{\{golden\_answers\}} \newline
Judgement: \\ 

% ================= Module 2 =================
\midrule
\multicolumn{2}{c}{\cellcolor{graybg}\textbf{Module: Knowledge Graph Refinement}} \\ 
\midrule

\textbf{System Prompt} & 
You are a professional AI assistant very good at refining the flawed knowledge graph. The knowledge graph is extracted from the documents for a given question so that it can be helpful for answering the question. But the current knowledge graph may contain incorrect information, redundant information, or lack critical information, making it impossible to deduce the correct answer for the given question. You will be given a question, a set of documents, the flawed knowledge graph and golden answers. Your task is to add, remove, or modify the content in the knowledge graph to make it accurate and relevant to answering the question. Make sure that the refined knowledge graph's content is both relevant to answering the question and entirely derived from the provided document.

\vspace{0.5em}
\textbf{Important Notes:}\newline
1. The new knowledge graph should be refined based on the flawed knowledge graph, \textbf{don't start from scratch}.\newline
2. The refined knowledge graph should be of the same format as the flawed knowledge graph.\newline
3. \textbf{Do not directly add the golden answers to the knowledge graph}, the refined knowledge graph should be derived from the documents.\newline
4. Do not output the explanation of your changes, only output the refined knowledge graph! \\ 
\cmidrule(l){2-2}
\textbf{User Prompt} & 
Question: \texttt{\{question\}} \newline
Documents: \texttt{\{reference\}} \newline
Flawed Knowledge Graph: \texttt{\{$s^-$\}} \newline
Golden Answers: \texttt{\{golden\_answers\}} \newline
Refined Knowledge Graph: \\ 
\bottomrule

\end{tabularx}
\caption{Prompts used for Data Construction and Refinement.}
\label{tab:prompt-data-construct}
\end{table*}

\end{document}